\documentclass[conference]{IEEEtran}
\IEEEoverridecommandlockouts

\usepackage{cite}
\usepackage{amsmath,amssymb,amsfonts}
\usepackage{algorithmic}
\usepackage{graphicx}
\usepackage{textcomp}
\usepackage{xcolor}
\usepackage{algorithm}
\usepackage{algorithmic}
\usepackage{booktabs} 
\usepackage{multirow}
\usepackage{amssymb}
\usepackage{longtable}
\usepackage{color}
\usepackage{amsmath}
\usepackage{mathrsfs}

\usepackage{subcaption}
\usepackage{longtable}
\usepackage{ltxtable}
\usepackage{tabularx}
\usepackage{amsmath}
\usepackage{mathrsfs}
\usepackage{multicol}
\usepackage{lipsum}
\usepackage{booktabs}
\usepackage{multirow}
\usepackage{dsfont}
\usepackage{mathbbol}
\usepackage{cleveref}
\usepackage{bbm}
\usepackage{url}
\usepackage{graphicx}
\usepackage{bbding}
\usepackage{newtxtext,newtxmath} 

\def\BibTeX{{\rm B\kern-.05em{\sc i\kern-.025em b}\kern-.08em
    T\kern-.1667em\lower.7ex\hbox{E}\kern-.125emX}}
\begin{document}

\title{AimTS: Augmented Series and Image Contrastive Learning for Time Series Classification}

\author{\IEEEauthorblockN{Yuxuan Chen\textsuperscript{1}$^{*}$, Shanshan Huang\textsuperscript{1}$^{*}$, Yunyao Cheng\textsuperscript{2}, Peng Chen\textsuperscript{1}, Zhongwen Rao\textsuperscript{3}, Yang Shu\textsuperscript{1}\textsuperscript{\Envelope},\\ Bin Yang\textsuperscript{1}, Lujia Pan\textsuperscript{3}, Chenjuan Guo\textsuperscript{1}}
\IEEEauthorblockA{\textit{\textsuperscript{1}East China Normal University, \textsuperscript{2}Aalborg University, \textsuperscript{3}Huawei Noah’s Ark Lab} \\
\{chenyx\_50,ssh,pchen\}@stu.ecnu.edu.cn,\\
\{yshu,byang,cjguo\}@dase.ecnu.edu.cn,\\
\{yunyaoc\}@cs.aau.dk,\{raozhongwen,panlujia\}@huawei.com}
\thanks{$^{*}$Equal contribution.}
\thanks{\textsuperscript{\Envelope}Corresponding author.}
}

\maketitle

\begin{abstract}
Time series classification (TSC) is an important task in time series analysis. Existing TSC methods mainly train on each single domain separately, suffering from a degradation in accuracy when the samples for training are insufficient in certain domains. The pre-training and fine-tuning paradigm provides a promising direction for solving this problem. However, time series from different domains are substantially divergent, which challenges the effective pre-training on multi-source data and the generalization ability of pre-trained models. To handle this issue, we introduce Augmented Series and Image Contrastive Learning for Time Series Classification (AimTS), a pre-training framework that learns generalizable representations from multi-source time series data. We propose a two-level prototype-based contrastive learning method to effectively utilize various augmentations in multi-source pre-training, which learns representations for TSC that can be generalized to different domains. In addition, considering augmentations within the single time series modality are insufficient to fully address classification problems with distribution shift, we introduce the image modality to supplement structural information and establish a series-image contrastive learning to improve the generalization of the learned representations for TSC tasks. Extensive experiments show that after multi-source pre-training, AimTS achieves good generalization performance, enabling efficient learning and even few-shot learning on various downstream TSC datasets. 

\end{abstract}



\section{Introduction}

\begin{figure}[h]
    \centering
\includegraphics[width=0.45\textwidth]{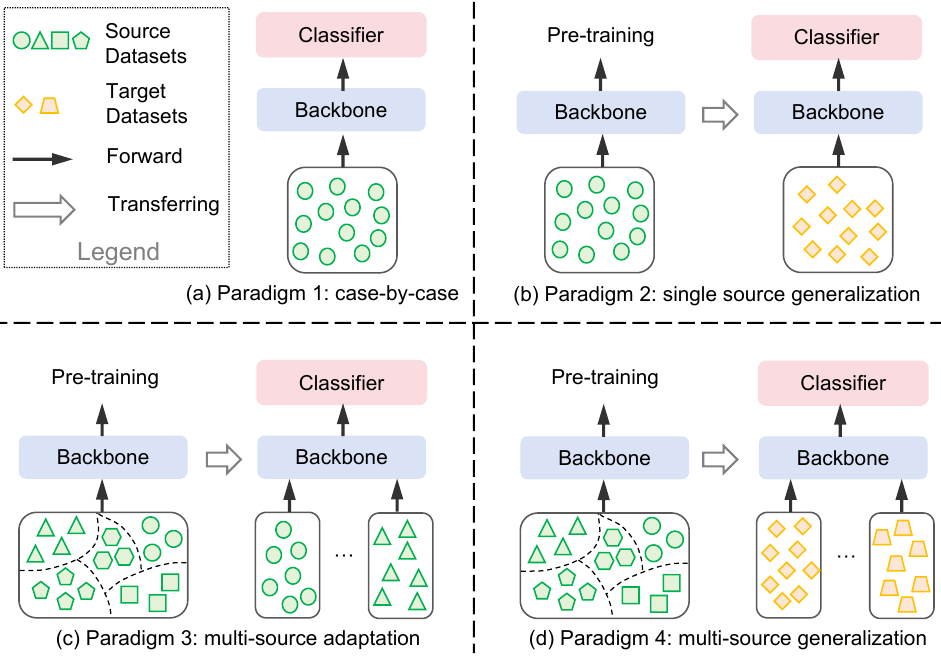}
    \caption{Illustration of existing deep learning methods for TSC.}
    \label{fig:intro1}
\end{figure}

Time Series Classification (TSC) is a classical and challenging task in many domains, such as action recognition \cite{icde1}, healthcare \cite{peng2017acts}, and transportation \cite{icde2}.
Ensuring high classification accuracy requires a large number of training samples, especially for recent deep  models.
As shown in Fig. \ref{fig:intro1}, existing time series classification methods mainly follow three paradigms: (a) the case-by-case paradigm \cite{rocket, timesnet, TS2Vec}, where a specific model is trained for each dataset and tested on the same dataset, (b) the single source generalization paradigm \cite{tfc, sim-mtm}, where a transferable model is trained on one dataset and then transferred to another dataset for fine-tuning and inference, and (c) the multi-source adaptation paradigm \cite{units, moment}, where a general model is pre-trained on multiple datasets that consist of downstream datasets.

However, obtaining enough training data for each task may not always be practical, as labeling time series data is inherently challenging and requires considerable expertise. 
For example, interpreting an Epilepsy series to assess health is difficult for most people, and only medical professionals can reliably label it as healthy or unhealthy, resulting in the insufficiency of training samples.
Using Paradigm 1 on such datasets with scarce training samples is highly prone to overfitting. 
Paradigm 2 aims to perform pre-training using single-source data and then transfer it to downstream data to overcome the limitation of data. 
However, this approach performs poorly when there is a significant domain difference between the pre-training and downstream data, such as between Gesture and Epilepsy.
Paradigm 3 uses multi-source data for pre-training, but its performance is less effective when the downstream data, such as the Epilepsy series, are not available in the pre-training dataset.
This motivates us to design a self-supervised pre-training method that learns general representations from multiple data sources and then fine-tunes the model using a few samples at specific downstream tasks that are not necessarily seen during pre-training (Fig. 1(d)). 
We thus propose a novel multi-source generalization paradigm to address the issue of data limitations.
Pre-training on multi-source data, compared to single-dataset pre-training, helps the model learn more diverse time series patterns. 
Meanwhile, self-supervised learning addresses the issue caused by insufficient training data, thus effectively solving the label scarcity. 
Representations obtained in this paradigm exhibit stronger universality, leading to better performance on downstream tasks in various domains. 

Existing self-supervised methods, such as contrastive learning methods~\cite{simclr,tfc,ts-tcc}, have demonstrated their effectiveness on classification tasks. These methods are supported by various data augmentation strategies~\cite{augmentation1,augmentation2,augmentationimage1} that treat augmentations with the same sample as positive pairs and with the other samples as negative pairs to improve the accuracy and generalization of the models. However, time series data from different domains show significant divergence due to semantic shifts~\cite{augmentation1,qiu1,qiu2,qiu3,qiu4}. 
This makes it hard for existing contrastive learning and data augmentation methods to generalize across multiple domain data in pre-training. 



\begin{figure}
    \centering
    \includegraphics[width=0.9\linewidth]{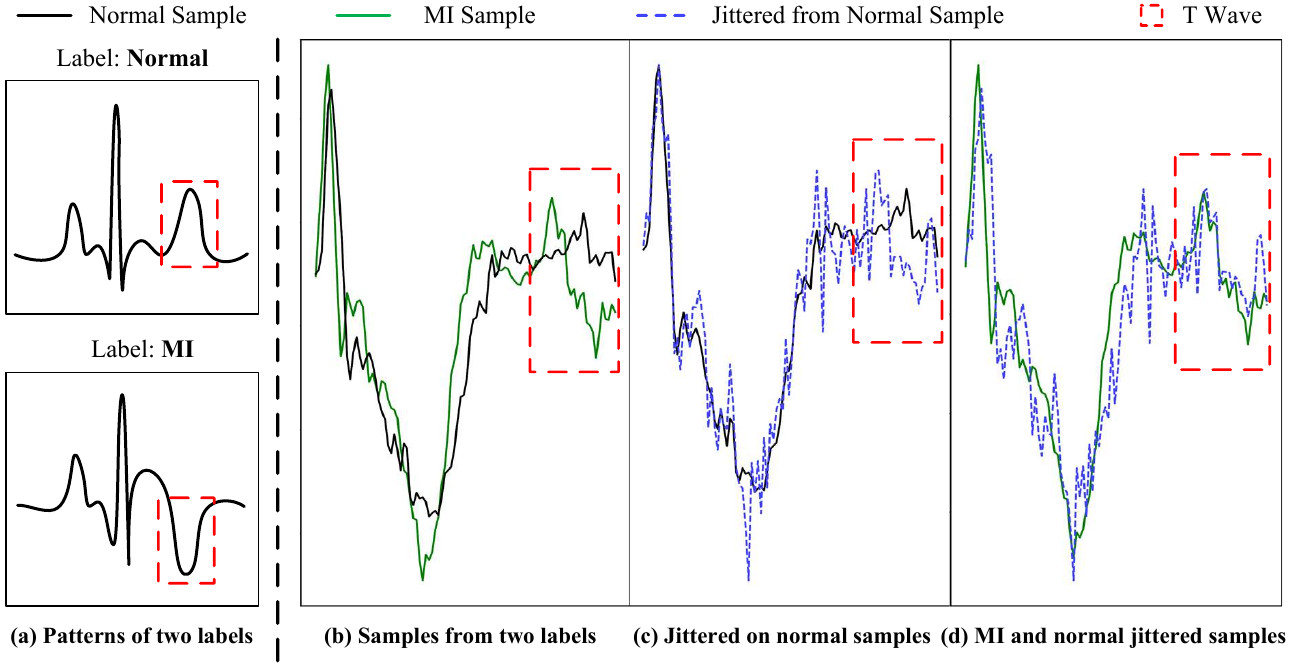}
    \caption{The example of augmentation causing semantic changes. The ECG200 dataset records electrocardiograms for healthy and myocardial infarction (MI) patients. (a) Pattern illustrations of two labels. T wave inversion is a sign of MI. 
    (b) The black line shows a normal ECG and the green line shows an MI ECG. (c) The blue dashed line shows jitter augmentation on the normal ECG and the T wave of this augmented sample has been inverted. (d) After jittering, the normal sample becomes more similar to the MI sample, leading to a change in its semantics.}
    \label{fig:intro2}
\end{figure}

\textbf{The first challenge is that }different data augmentations may fail to maintain the same semantics of the original time series, making contrastive learning with augmented time series challenging in multiple domain pre-training.
Existing methods \cite{ts-tcc,tfc} treat various augmented views of the same time series samples as positive pairs when conducting contrastive learning, based on the assumption that the augmented data retains semantic information similar to the original data, which is key for distinguishing different samples. However, some data augmentations may change the semantics of the original samples \cite{augmentation2}.
For example, applying jitter augmentations to a motion time series may not change the motion state, but applying it to an Electrocardiogram (ECG) time series may cause it to shift from healthy to unhealthy \cite{augmentation1} as shown in Fig. \ref{fig:intro2}. Bringing together two semantically different views as positive pairs confuses the model and thus influences the discrimination performance of the learned representation \cite{wang2023sncse}. When facing multiple domain data, we are unable to manually investigate whether the augmented samples exhibit the same semantics as the original time series. As a result, effectively leveraging multiple augmentations while avoiding incorrect use of augmented data is challenging.

\textbf{The second challenge is that} data augmentations within the single time series modality restrict the model's ability to learn general representations from the entire time series samples across multiple domains.
Morphological (i.e., structural) information, such as the composition of lines or curves is crucial for distinguishing categories in TSC~\cite{tsec,shapelet}, as shown in Fig \ref{fig:intro2}. 
However, the time series modality describes the data as a sequence of values changing over time. 
It mainly captures statistical information based on numerical values which may still be limited in solving 
classification problems with distribution shifts across datasets. 
Merely augmenting time series data does not learn to fully capture structural information which helps generalization for TSC in a complementary way.
To solve these problems, we propose an \textbf{A}ugmented Series and \textbf{Im}age Contrastive Learning for \textbf{T}ime \textbf{S}eries Classification (AimTS), which learns generalizable representations by augmenting from both time series values and structures for TSC in multi-source pre-training.

\textbf{To address the first challenge}, we propose a two-level prototype-based contrastive learning, including inter-prototype contrastive learning and intra-prototype contrastive learning.
Since most augmentations do not change the semantics of the original sample, aggregating augmented samples into a prototype minimizes the influence of semantic changes that may be caused by some augmentation.
Different from existing prototypes aggregated from the same class samples \cite{prototypicalcon}, we propose novel inter-prototype contrastive learning where prototypes are learned from multiple augmented samples.
In inter-prototype contrastive learning, the prototypes of different samples serve as negative pairs, while the prototype and its corresponding sample serve as the positive pair. 
When training with multi-source data, augmentations influence different domains differently. 
To further enable the generalization of learned representations, different augmentation methods should contribute equally to the prototype. 
Thus, we propose intra-prototype contrastive learning across augmentations with an adaptive temperature. 
It encourages a uniform distribution of representations from different augmentations, allowing the aggregated prototypes to make full use of all augmentations and not be dominated by specific ones.

\textbf{To address the second challenge}, we propose series-image contrastive learning with the purpose of learning general time series representations by simultaneously capturing the numerical and structural information from both time series and image modalities.
We first convert each time series sample into an RGB image. Different from existing modeling on image solely, we propose to encode the image and time series separately to extract representations of each modality.
Next, the series-image contrastive learning treats the corresponding image of each time series sample as the positive sample, and treats images from other samples as its negative samples. 
Simply using the images as negative samples is not sufficient to distinguish different time series samples, because their numerical aspects are missing, which is also crucial in TSC.
We further design a novel geodesic series-image mixup strategy to create mixed-modality representations as negative samples that consider both numerical and structural aspects of time series, thereby better distinguishing time series samples that belong to different classes.

In summary, our contributions are as follows:

\begin{itemize}
    \item We propose the first TSC pre-training framework to learn general time series representations from multiple datasets that improve performance in various downstream datasets.
    \item We design a prototype-based contrastive learning method that effectively augments multi-source datasets during pre-training to achieve generalized representations.
    \item We introduce image modality to overcome the limitations of single-modality augmentation strategies and leverage the image modality for more generalizable representations with series-image contrastive learning.
    \item Extensive experiments show that AimTS achieves good generalization performance for downstream classification tasks with an average accuracy of 0.870 on the 128 UCR datasets and 0.780 on the 30 UEA datasets and outperforms the state-of-the-art methods.
\end{itemize}

\begin{figure*}[t]
    \centering
    \includegraphics[width=0.95\textwidth]{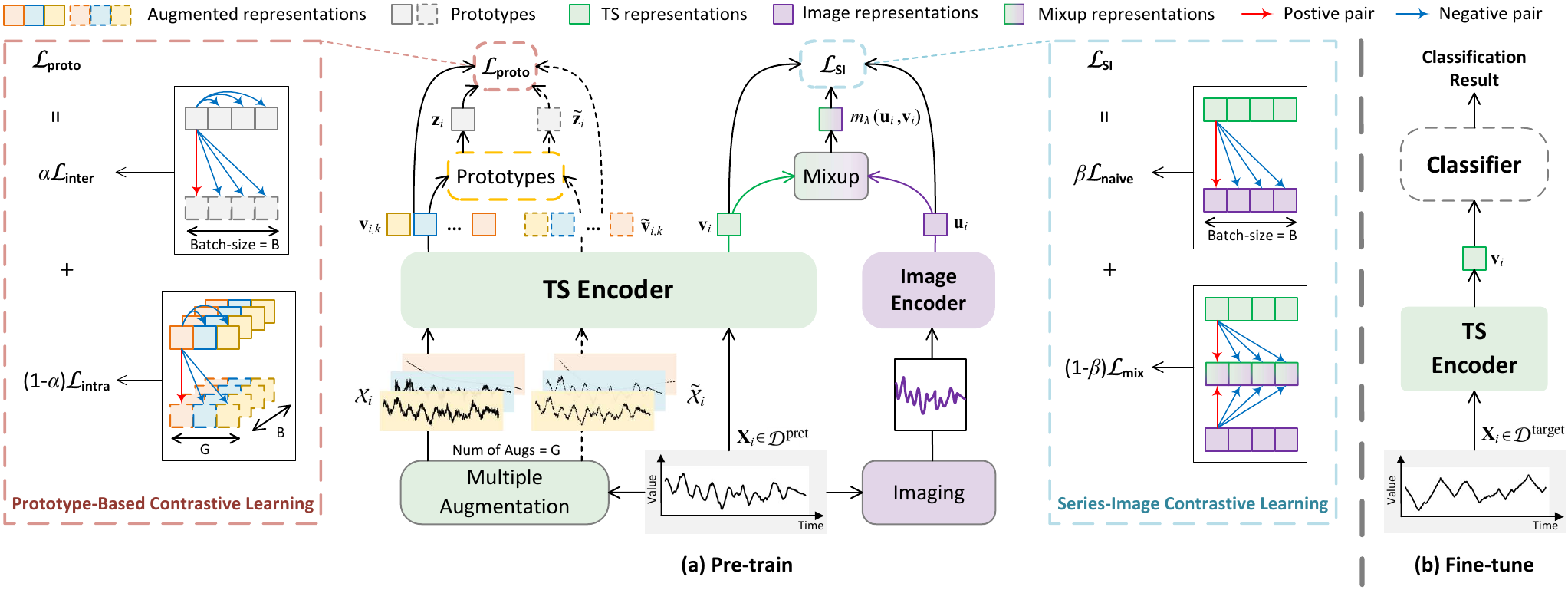}
    \caption{Overview of AimTS.}
    \label{fig:F1}
\end{figure*}

\section{Related Work}
\subsection{Contrastive Learning for Time Series} 
Contrastive learning, as a common pre-training method, has achieved success in many areas~\cite{contrastive1,qiu5,qiu6}. For time series analysis, such unsupervised representation learning methods have achieved good performance in various tasks. T-Loss~\cite{tloss} uses a random subseries from a time series and treats them as positive pairs when they belong to the subseries, and negatives if belong to the subseries of other series. TNC~\cite{tnc} defines the temporal neighborhood of windows using a normal distribution, treating samples within the neighborhood as positives and those outside as negatives. TS-TCC~\cite{ts-tcc} uses weak and strong augmentations to generate two views of data. TS2Vec~\cite{TS2Vec} proposes augmented context views to obtain representations of various semantic levels of time series. TimesURL~\cite{timesurl} proposes double universums for constructing negative pairs and introduces time reconstruction. CoST~\cite{cost} introduces a frequency-domain contrastive loss to learn disentangled trend and seasonal representations separately. TFC~\cite{tfc} proposes a contrastive learning objective of minimizing the distance between time-based and frequency-based embeddings. SoftCLT~\cite{softcl} introduces soft assignments ranging from 0 to 1 for contrastive losses. Unlike these methods, AimTS no longer limit to extracting representations in specific datasets and obtains generalized representations for downstream time series classification tasks through multi-source pre-training.

\subsection{Adaptive Data Augmentation}
Data augmentation is a key component of contrastive learning. Contrastive learning through augmentations of different samples has been applied across various fields of deep learning~\cite{simclr,moco,clear}. Research across various fields has shown that the most suitable augmentations vary depending on the target task and dataset~\cite{augmentation1,augmentation2}. In the field of time series, there have been studies focused on developing methods to adaptively select the optimal augmentations and parameters for a given dataset. CADDA~\cite{CADDA} proposes a gradient-based framework that extends the bilevel framework of AutoAugment \cite{autoaugment} to search class-wise data augmentation policies for EEG signals. InfoTS~\cite{infots} proposes a criterion for selecting effective augmentations based on information-aware definitions of high fidelity and diversity. AutoTCL~\cite{autocl} proposes a factorization-based adaptive framework for searching data augmentations which summarizes the most commonly used augmentations in a unified form and extends them into a parameterized augmentation approach. Although these studies enable adaptive search in contrastive learning, they are limited to single-dataset applications and cannot simultaneously select the optimal augmentations and parameters for multiple target datasets.

\subsection{Image Modality on Time Series}
Using the image modality for time series analysis is an underexplored field. Existing methods visualize time series data through methods such as Gramian fields \cite{GAF}, recurrence plots \cite{Hatami2017ClassificationOT, Langley2000CraftingPO}, and Markov transition fields \cite{Wang2015SpatiallyET}. These approaches require domain experts to design specialized imaging techniques, which are not universally applicable. ViTST \cite{li2023time} plots time series as line charts and achieves promising results, suggesting that extensive specialized designs may not be necessary for effective visualization. Apart from time series classification tasks, recent works have also explored using images for time series forecasting and anomaly detection. VisionTS~\cite{visionts} converts time series into binary images for forecasting, while HCR-AdaAD~\cite{hcr-adaad} extracts representations from time series images to aid in anomaly detection. However, current methods often discard the original time series data after converting them to images, focusing only on image analysis. AimTS addresses this limitation by simultaneously handling both time series and image modalities, enhancing the performance of TSC tasks through the integration of image modality modeling.


\section{Preliminaries}
We first cover important concepts and then present the problem statement.

\subsection{Definitions}
\textit{Definition 1:} \textbf{Time Series.} A time series is defined as $ \mathbf{X} = \langle\mathbf{x}_1, \mathbf{x}_2, \dots, \mathbf{x}_M\rangle\in \mathbb{R}^{M\times T} $, where $M$  is the number of variables (or dimensions) and  $T$  is the number of time steps. When  $M=1$, it is an univariate time series. When  $M>1$, it is a multivariate time series and we also refer to $\mathbf{X}$ as a time series sample.

 \textit{Definition 2:} \textbf{Time Series Classification.} Time series classification is the task of assigning a predefined class label to a time series. Given a dataset $\mathcal{D} = \{\mathbf{X}_1, \mathbf{X}_2, \dots, \mathbf{X}_n\}$, where each  $\mathbf{X}_i \in \mathbb{R}^{M \times T}$ and $n$  is the number of samples, the goal is to learn a mapping function  $f(\mathbf{X}_i) \rightarrow y_i$  that assigns each time series  $\mathbf{X}_i$  to a label  $y_i \in \{1, 2, \dots, C\}$, where  $C$  is the number of classes.

\textit{Definition 3:} \textbf{Augmented View.} Data augmentation $g(\cdot)$  is a technique used to artificially expand a dataset by modifying real data samples. An augmented view of a time series sample $\mathbf{X}$ is a transformed time series  $\mathbf{X}^{\prime} = [g(\mathbf{x}_1), g(\mathbf{x}_2), \dots, g(\mathbf{x}_M)] $, where  $g(\cdot)$ is applied to each variable $\mathbf{x}\in\mathbb{R}^{T}$. 

\textit{Definition 4:} \textbf{Contrastive Learning.} Contrastive learning aims to learn representations by bringing similar pairs closer and pushing dissimilar pairs apart. Given an anchor $\mathbf{X}$, a positive view $\mathbf{X}^+ $, and negative views  $\mathbf{X}^-$ , the contrastive loss is defined as:
\begin{equation*}
\mathcal{L} = -\log \frac{\exp(\mathbf{r}\cdot \mathbf{r}^+) }{\exp(\mathbf{r}\cdot \mathbf{r}^+ ) + \sum_{\mathbf{r}^-} \exp(\mathbf{r}\cdot \mathbf{r}^-) },
\end{equation*}
where  $\mathbf{r}$, $\mathbf{r}^+$ and $\mathbf{r}^-$ are the representations of $\mathbf{X}$, $\mathbf{X}^+$ and $\mathbf{X}^-$, respectively.

\subsection{Problem Statement}
During pre-training, we use a dataset composed of $K$ different subdatasets or source domains $\mathcal{D}^{\mathrm{pret}}=\bigcup_{k=1}^K\mathcal{D}^{\mathrm{pret}_k}$. The $k$-th source domain is represented as $\mathcal{D}^{\mathrm{pret}_k}=\{\mathbf{X}_{i}^{k}\mid i=1,\ldots,N_k\}$, where $N_k$ is the number of samples in this resource and $N=|\mathcal{D^\mathrm{pret}}|=\sum_{k=1}^KN_k$ is the overall number of time series samples for pre-training. The $i$-th sample in the $k$-th source domain is represented as $\mathbf{X}_{i}^{k}\in\mathbb{R}^{M_k \times T_k}$ where $M_k$ is the number of its variables and $T_k$ is its length. The goal of pre-training is to learn a model $F(\cdot)$ to obtain generalizable representations from $\mathcal{D}^{\mathrm{pret}}$, which can help the classification tasks on new domains. We denote $\mathcal{D}^{\mathrm{target}}$ as one of the target datasets for downstream classification, and $\mathcal{D}^{\mathrm{target}}_\mathrm{train}= \{(\mathbf{X}_{i}, y_{i})\mid i=1,\ldots, N_\mathrm{target} \}(N_\mathrm{target}<<N)$ as its training data, where $y_{i}$ is the label of the time series $\mathbf{X}_{i}\in \mathbb{R}^{M \times T}$ and $N_\mathrm{target}$ represents the number of training samples in this downstream task. This training set is used to fine-tune the pre-trained model $F(\cdot)$ and train the task-specific classifier $P^\mathrm{cls}$, which will then make accurate classifications for each target data $\mathbf{X}_i\in\mathcal{D}^{\mathrm{target}}_\mathrm{test}$ as $\hat{\mathbf{y}}_i=P^\mathrm{cls}(F(\mathbf{X }_i))$.

\begin{figure*}[t]
    \centering
\includegraphics[width=0.95\textwidth]{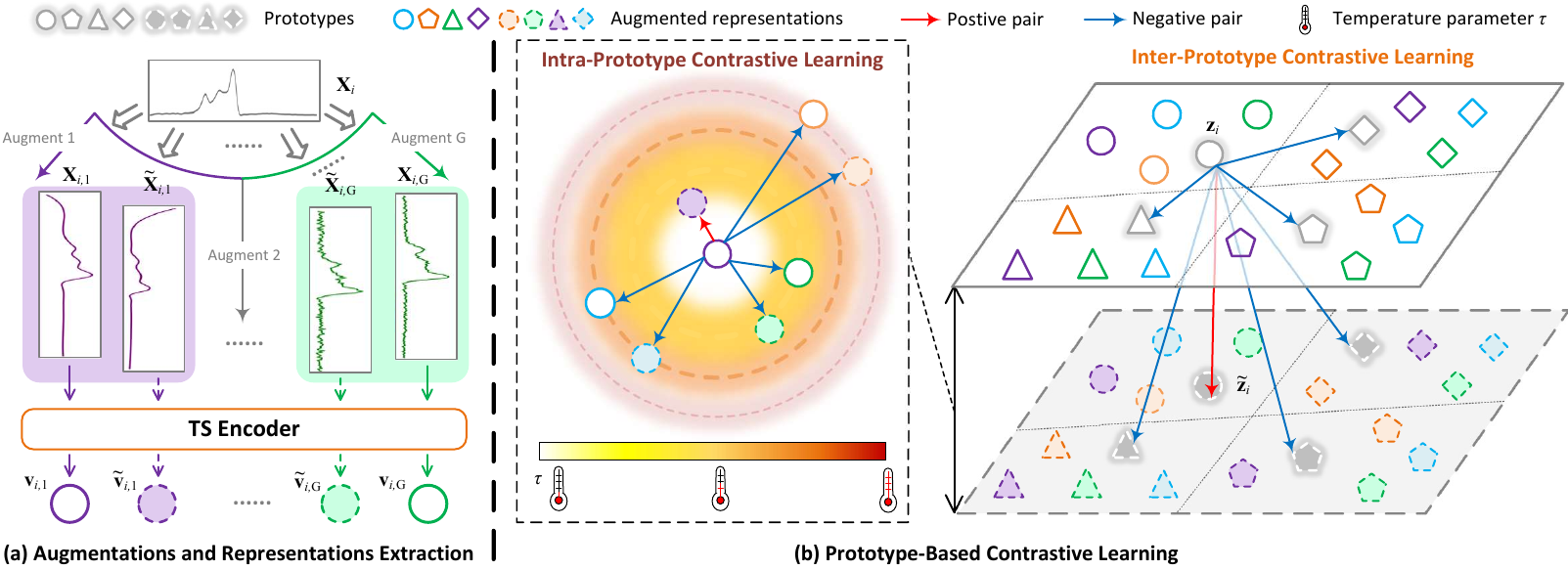}
    \caption{Overview of prototype-based contrastive learning. Solid and dashed shapes of the same color and the same shape represent the representations of the two augmented views from the same augmentation on the same sample.  Different colors indicate different augmentation methods. Different shapes represent distinct samples. Gray shapes denote prototypes for each sample, with two prototypes generated per sample. (a) Perform G types of augmentations on time series data by applying each twice to extract representations.  (b) Illustration of prototype-based contrastive learning. The \textit{intra-prototype contrastive learning} is conducted within the same sample and its augmentations. The color band indicates the distance between representations of different augmentations, where transitioning from white to red indicates the increase of their distance, and the decrease of the temperature parameter $\tau$ accordingly. The \textit{inter-prototype contrastive learning} is conducted across different samples. 
    }
    \label{fig:MU}
\end{figure*}

\section{Methodology} \label{Methodology}




\subsection{Overall Framework.}
We propose AimTS, a pre-training framework designed to conduct TSC tasks by enhancing the generalization of representations from multi-source datasets through prototype-based and series-image contrastive learning. Fig. \ref{fig:F1} gives an overview of the AimTS framework, which consists of a pre-training stage and a fine-tuning stage. We first pre-train a time series (TS) encoder and an image encoder using two contrastive learning tasks as shown in Fig. \ref{fig:F1}(a), and then transfer the pre-trained TS encoder via fine-tuning in a downstream task and train a classifier as shown in Fig. \ref{fig:F1}(b).

\textbf{At the pre-training stage}, for each input time series sample, we apply multiple data augmentations to generate several augmented views. These views are then fed into TS encoder to produce their respective representations. By aggregating the representations of these different views, we obtain a prototype for the sample. Based on these prototypes, we propose intra-prototype contrastive learning and inter-prototype contrastive learning. By adding these two contrastive strategies, we propose a two-level prototype-based loss $\mathcal{L}_{\mathrm{proto}}$ to capture generalized representations.

Meanwhile, each time series sample is converted into an image.
The image encoder takes as input the standardized image generated from the time series data and converts the input into an image representation. To supplement general time series representations learning, we conduct series-image contrastive learning loss $\mathcal{L}_{\mathrm{SI}}$ between the representations of the time series and the corresponding image representations with the geodesic mixup strategy.

During pre-training, AimTS optimizes the parameters of the TS encoder and the image encoder through the prototype-based contrastive loss $\mathcal{L}_{\mathrm{proto}}$ and the series-image contrastive loss $\mathcal{L}_{\mathrm{SI}}$. The overall loss is defined as:
\begin{equation}
\mathcal{L}=\mathcal{L}_{\mathrm{proto}}+\mathcal{L}_{\mathrm{SI}}.
\end{equation}
 
\textbf{At the fine-tuning stage}, we use data of training set from a target dataset to fine-tune the parameters of the pre-trained TS encoder and train a task-specific classifier. At this stage, the time series input is directly fed into the TS encoder without any data augmentation or conversion into images, producing a representation. This representation is then passed to a classifier to obtain a probability distribution over the class variable values. Using the cross-entropy loss, we fine-tune the parameters of the TS encoder while training the task-specific classifier for the downstream task.

\subsection{Prototype-Based Contrastive Learning}
To ensure the effective utilization of various data augmentations in multi-source pre-training to obtain generic representations that could be applied to different downstream classification tasks, we propose a novel two-level prototype-based contrastive learning as the first learning objective of the framework. Here, we first detail the generation of prototypes by aggregating multiple augmented views and then construct prototype-based contrastive learning, including intra-prototype contrastive learning and inter-prototype contrastive learning.

\subsubsection{Prototype generation}
Views generated by different data augmentation operations capture the characteristics of time series data under various transformations. However, for a given dataset, it is often unclear which data augmentation methods may distort the semantics of the data in the context of multiple domain pre-training. 
Considering that most augmentations do not alter the semantics of the original sample as used by the existing methods~\cite{augmentation1,augmentation2}, we aggregate augmented views of a time series sample into a prototype to minimize the potential negative impact of any semantic changes introduced by specific augmentations. 
Meanwhile, aggregating augmented views can produce more stable sample representations, reducing randomness or noise within the representations and highlighting the essential characteristics of the original data.

As shown in Fig. \ref{fig:MU}(a), for each time series sample $\mathbf{X}_{i}$, we first randomly generate two different augmented views using each augmentation from a data augmentation bank that contains $G$ types of augmentations, meaning that we generate two sets of augmented views 
$\mathcal{X}_{i} = \{ \mathbf{X}_{i,1},\mathbf{X}_{i,2}, \cdots, \mathbf{X}_{i,G} \} $ and $\tilde{\mathcal{X}}_{i} = \{ \tilde{\mathbf{X}}_{i,1},\tilde{\mathbf{X}}_{i,2}, \cdots, \tilde{\mathbf{X}}_{i,G}\}$, where $\mathbf{X}_{i,k}$ and $\tilde{\mathbf{X}}_{i,k}$ represent the two augmented views of the $i$-th sample $\mathbf{X}_{i}$ obtained by the $k$-th augmentation method using different randomized parameters, respectively. Then, augmented views $\mathbf{X}_{i,k}$, $\tilde{\mathbf{X}}_{i,k}$ are fed into a TS encoder \( F^{\mathrm{TS}}(\cdot) \) to obtain their high-dimensional latent representations $\mathbf{r}_{i,k}$, $\tilde{\mathbf{r}}_{i,k}$.
Finally, we use the average of the representations of various augmented views as the prototype. 
The prototype of $\mathbf{X}_{i}$ is formulated as:
\begin{equation}
\textbf{z}_{i}=P^{\mathrm{TS}}\bigl(\frac{1}{G}\sum_{k=1}^{G}( (\mathbf{r}_{i,k})\bigr), 
\tilde{\mathbf{z}}_{i}=P^{\mathrm{TS}}\bigl(\frac{1}{G}\sum_{k=1}^{G}( (\tilde{\mathbf{r}}_{i,k})\bigr),
\end{equation}
where $\mathbf{z}_{i}, \tilde{\mathbf{z}}_{i}\in \mathbb{R}^J$ and $G$ denotes the number of augmentations.


An augmented view whose amplitude is significantly different from others may dominate the representation distribution of the prototype. 
This view can significantly affect the value of the prototype. Ideally, the prototype should balance the representation of all views rather than being dominated by any single view. 
Thus, in contrastive learning, this requires additional handling to prevent a single view’s representation from dominating the prototype, ensuring a fair contribution from each view.

\subsubsection{Intra-prototype contrastive loss}

To achieve a uniform distribution of different augmented views within the representation space, we propose intra-prototype contrastive learning with adaptive temperature parameters. We use a temperature parameter $\tau$ to control the strength of penalties on negative samples. Specifically, a lower-temperature contrastive loss imposes greater penalties on negative samples, leading to more separated representations. We design different $\tau$ for each negative pair in intra-prototype contrastive loss.

For $\mathbf{X}_{i}$, views generated from the same $k$-th augmentation, $\mathbf{X}_{i,k}$ and $\tilde{\mathbf{X}}_{i,k}$, are treated as a positive pair (e.g., the purple solid circle and the purple dashed circle in Fig. \ref{fig:MU}(b)). Views generated from different augmentations, such as $\mathbf{X}_{i,j}$ and $\mathbf{X}_{i,k}$, are treated as negative pairs (e.g., the purple solid circle and circles of other colors in Fig. \ref{fig:MU}(b)). Then, we change $\tau$ for each negative pair to control the separation of different augmented views in the representation space. 
Specifically, for two views with greater distance, we increase $\tau$ to make their representations in the representation space more similar (e.g., the purple solid circle and blue circle). Conversely, for views that are already similar (with smaller distances), we reduce $\tau$ to make their representations better distinguished within the space (e.g., the purple solid circle and green circle).


To obtain $\tau$ for each pair, we first use a distance metric $D(\text{·},\text{·})$ to obtain the distance $d_{i}^{(j,k)}=D(\mathbf{X}_{i,j},\mathbf{X}_{i,k})$ between $\mathbf{X}_{i,j}$ and $\mathbf{X}_{i,k}$, which are originated from the $j$-th and $k$-th types of augmentation, respectively. Then, we use the softmax function to map the distances to $\tau$. The $\tau$ for a pair of $\mathbf{X}_{i,j}$ and $\mathbf{X}_{i,k}$ is formulated as:
\begin{equation}
\label{eq:augloss}
\tau_{i}^{(j,k)}=\tau_0+\frac{\exp(d_{i}^{(j,k)})}{\sum_{k=1}^{G} \exp(d_{i}^{(j,k)})}.
\end{equation}


Performing contrastive learning among multiple augmented views within a prototype helps prevent views with significantly different amplitude from dominating the prototype, thereby indirectly optimizing the aggregation from views to the prototype. To output a lower-dimensional representation for contrastive learning, $\mathbf{r}_{i,k}$ and $\tilde{\mathbf{r}}_{i,k}$ are input into a non-linear projection \( P^{\mathrm{TS}}(\cdot) \) to output the low-dimensional representations $\mathbf{v}_{i,k}=P^{\mathrm{TS}}(\mathbf{r}_{i,k})$ and $\mathbf{\tilde{v}}_{i,k}=P^{\mathrm{TS}}(\mathbf{\tilde{r}}_{i,k})$. 
The intra-prototype contrastive loss for $\mathbf{X}_{i}$ is defined as:
\begin{equation}
    \ell_{i}^{\mathrm{intra}}=-\sum_{k=1}^{G}\log\frac{\exp(\tilde{s}_{i}^{(k,k)})}{\sum_{j=1}^{G}\bigl(\mathbbm{1}_{[k\neq j]}\exp(s_{i}^{(k,j)})+\exp(\tilde{s}_{i}^{(k,j)})\bigr)},
\end{equation}
where $s_{i}^{(k,j)}=\mathbf{v}_{i,k}\cdot\mathbf{v}_{i,j}/\tau_{i}^{(k,j)}$ and $ \tilde{s}_{i}^{(k,j)}=\mathbf{v}_{i,k}\cdot\tilde{\mathbf{v}}_{i,j}/\tilde{\tau}_{i}^{(k,j)}$. Before calculating $\tau$, we set $d_{i}^{(j,j)}$to negative infinity so that $\tau_{i}^{(j,j)}=\tau_0$ to ensure that positive pairs are close to each other. 

\subsubsection{Inter-prototype contrastive loss} We select positive and negative pairs between different prototypes to identify discriminative information of samples. This discriminative information better captures the differences between samples, significantly improving classification performance in downstream tasks.
The $i$-th sample, 
$\mathbf{z}_{i}$ and $\tilde{\mathbf{z}}_{i}$ treat each other as positive samples (e.g., the gray solid circle and gray dashed circle), while they consider the prototypes of other samples in the batch as negative samples (e.g., the gray circle and circles of other colors). The inter-prototype contrastive loss for $\mathbf{X}_{i}$ is defined as:
\begin{equation}
\label{eq:insloss}
    \ell_{i}^{\mathrm{inter}}=-\log\frac{\exp(\mathbf{z}_{i}\cdot\tilde{\mathbf{z}}_{i}/\tau)}{\sum_{j=1}^{B}\bigl(\mathbbm{1}_{[i\neq j]}\exp(\mathbf{z}_{i}\cdot\mathbf{z}_{j}/\tau)+\exp(\mathbf{z}_{i}\cdot\tilde{\mathbf{z}}_{j}/\tau)\bigr)},
\end{equation}
where $B$ denotes the batch size.

With the above two training objectives, our prototype-based contrastive learning method effectively uses various augmentations to create generalized representations, improving the classification performance of downstream tasks. The overall prototype-based contrastive loss is defined as:
\begin{equation}
\mathcal{L}_{\mathrm{proto}}=\frac{1}{2B}\sum_{i=1}^{B}\bigl(\alpha\ell_{i}^{\mathrm{inter}}+(1-\alpha)\ell_{i}^{\mathrm{intra}}),
\end{equation}
where $\alpha$ are hyperparameters.
 
\subsection{Series-Image Contrastive Learning} \label{inter}
Augmentation from the time series modality that models from the numerical values makes it uneasy to capture the structural information, which is crucial for category recognition.
The time series data from different domains are always composed of lines or curve segments, and therefore, it is more straightforward to capture the structural information of time series based on the shapes than on the numerical values.
To capture the structural information of time series, we introduce image modality and propose series-image contrastive learning with a geodesic mixup strategy to capture the numerical and structural information simultaneously. 

First, we convert each time series sample into an image. 
Then, the time series sample and the corresponding image are treated as its positive pair, and the time series with images from other samples in the same batch serve as negative pairs. 
The naive contrastive learning establishes a correspondence between time series and images, but such differences remain insufficient to distinguish different time series samples, as the numerical aspects of time series are not considered in the negative samples.
Therefore, we designed a geodesic mixup to create mixed representations located between representations of these two modalities' subspaces.
Treat these mixed representations as negative for contrastive learning to expand the effective subspace of learned representations, making samples easier to classify.

\subsubsection{Image feature extraction}
Visualizing time series data through line charts is a natural intuition to translate numeric data into image modality. In a line chart, the x-axis means timestamps and the y-axis denotes values. 
We use the symbol ``*" to present the observed data points and connect them with straight lines. 
As shown in Fig. \ref{fig:F1}(a), for a multivariate time series sample $\textbf{X}_i$, we plot a line chart for each variable respectively, because each variable has a distinct scale, denoted as $\mathrm{Image}(\mathbf{X}_{i})$. 
We standardize images of different variables to the same square sizes. 
In addition, different colors are equipped for different variables, and the corresponding sub-image of each variable is stitched together into an image. 
We then use an image encoder to obtain the representations: $\mathbf{r}_{i}^{\mathrm{I}}=F^\mathrm{I}(\mathrm{Image}(\mathbf{X}_{i}))$. At the same time, we extract the corresponding representations $\mathbf{r}_i=F^{\mathrm{TS}}(\mathbf{X}_i)$ by the TS encoder.

\begin{figure}
    \centering
    \includegraphics[width=0.45\textwidth]{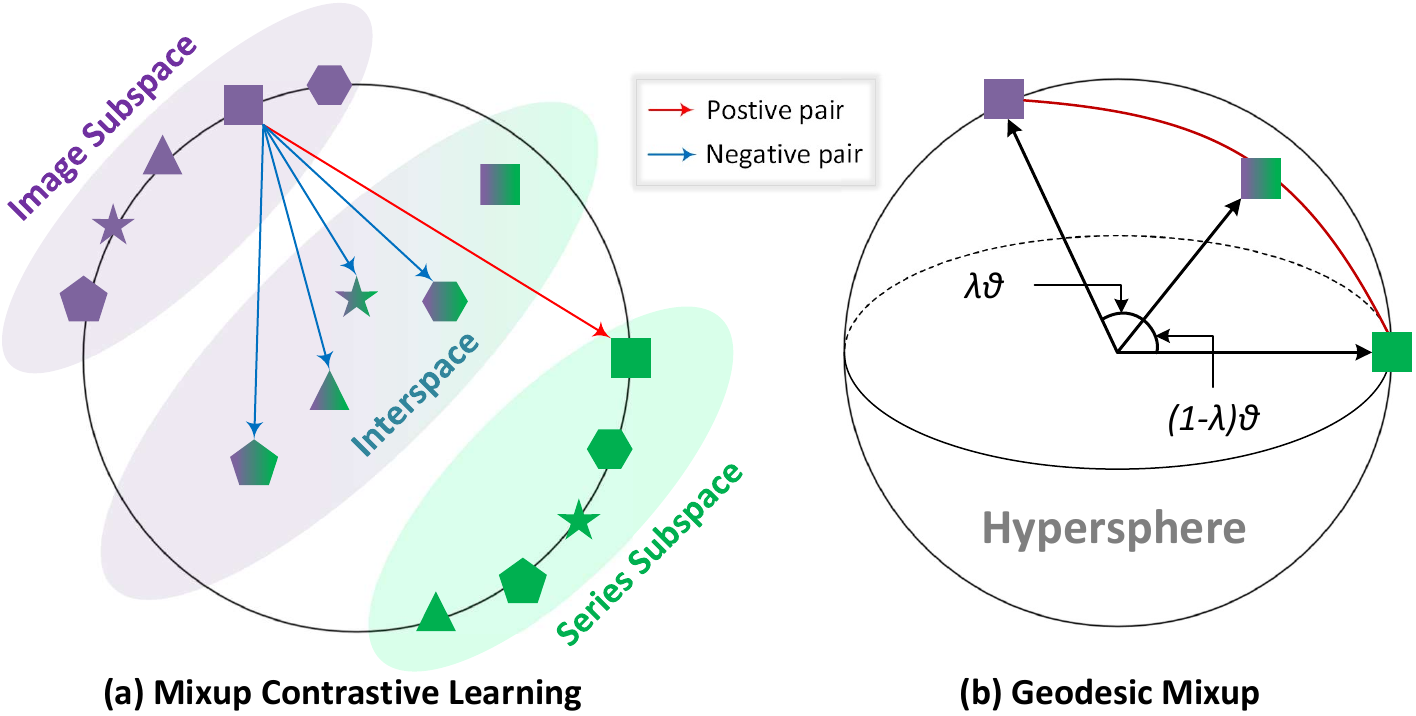}
    \caption{Geodesic mixup strategy and mixup contrastive learning. (a) Illustration of mixup contrastive learning. Various green shapes, such as stars, triangles, and others, represent the time series representations, while purple shapes indicate their corresponding image representations.  (b) Illustration of geodesic mixup. The green and purple squares represent the time series and image representations on the hyperspherical surface, respectively. The green-purple square represents the mixed representation by geodesic mixup, which can be seen to remain on the hypersphere.}
    \vspace{-0.4cm}
    \label{fig:mixup}
\end{figure}

\subsubsection{Series-image contrastive loss}
Time series modeling helps capture the numerical information of the data, while image modality modeling provides structural information. 
Capturing similar information between different modalities of each sample provides the unique aspect of each modality in learning general representation. 
Therefore, we present a series-image contrastive loss to maximize the similarity between TS representations and their image-based counterparts. It treats the input sample with the corresponding image as a positive pair, and the image generated by the other samples in the batch as the negative pairs. 

Due to the characteristics of different modalities, there exists incomparable information between the time series and image modalities. 
For instance, the contrast of an image is unrelated to the intrinsic properties of the time series data, such information is unsuitable for cross-modality contrastive learning.
Therefore, to filter out the incomparable information, we perform non-linear projections on the representations of each modality, allowing them to be compared during training.
We obtain more suitable time series representation $\mathbf{v}_i$ and image representation $\mathbf{u}_i$ from $\mathbf{r}_i$ and $\mathbf{r}_i^\mathrm{I}$ by filtering for series-image contrastive learning.
The series-image contrastive loss of the \textit{i}-th sample in a batch can be formulated as:
\begin{equation} \begin{split}     \ell_i^{\mathrm{I-S}}=-\mathrm{log}\frac{\exp\bigl(\text{sim}(\mathbf{u}_{i}\cdot\mathbf{v}_{i})/\tau\bigr)}{\sum_{j=1}^B\exp\bigl(\text{sim}(\mathbf{u}_{i}\cdot\mathbf{v}_{j})/\tau\bigr)}\\     \ell_{i}^{\mathrm{S-I}}=-\mathrm{log}\frac{\exp\bigl(\text{sim}(\mathbf{v}_{i}\cdot\mathbf{u}_{i})/\tau\bigr)}{\sum_{j=1}^{B}\exp\bigl(\text{sim}(\mathbf{v}_{i}\cdot\mathbf{u}_{j})/\tau\bigr)}, \end{split} \end{equation}  
where $\ell_i^{\mathrm{I-S}}$ is the $i$-th image representation contrast with all TS representations in one batch, and $\ell_{i}^{\mathrm{S-I}}$ is the $i$-th TS representation contrast with image representations. The naive series-image contrastive loss is defined as:
\begin{equation}    
\label{eq:si}
\mathcal{L}_{\mathrm{naive}}=\frac{1}{2B}\sum_{i=1}^{B}\left(\ell_{i}^{\mathrm{I-S}}+\ell_{i}^{\mathrm{S-I}}\right). \end{equation}
To this end, the series and image representations that locate in two subparts of the representation space are aligned.

\subsubsection{Geodesic mixup strategy} 
Although the series and image representations could be aligned by contrastive learning using Eq. \ref{eq:si}.
Past research~\cite{mindgap} has shown that, even in well-trained models, representations from the two modalities tend to be located in two separate subspaces of the whole representation space, as the image and the series subspace shown in Fig. \ref{fig:mixup}(a). 
This phenomenon means there is a large unexplored interspace between these two subspaces.
If representations of the two modalities appear in these interspaces, it indicates that the representation of one modality is closer to the representation of the other modality, and thus is more likely to contain information offered by the other modality, as the interspace shown in Fig. \ref{fig:mixup}(b).
This activates us to design a geodesic mixup strategy as:
\begin{equation}
\label{eq:mixup}
m_\lambda(\mathbf{u},\mathbf{v})=\mathbf{u}\frac{\sin(\lambda\theta)}{\sin(\theta)}+\mathbf{v}\frac{\sin((1-\lambda)\theta)}{\sin(\theta)},
\end{equation}
where $\theta=\cos^{-1}(\mathbf{u}\cdot\mathbf{v})$ is the angle between image representation $\mathbf{u}$ and series representation $\mathbf{v}$ measured by the geodesic distance, as shown by the red arc in Fig. \ref{eq:mixup}(b).
The parameter $\lambda\sim\mathrm{Beta}(\gamma,\gamma)$ is a random coefficient used to control the mixing ratio between the two representations and $\gamma$  is a hyperparameter. 
Supported by empirical work~\cite{liu2017sphereface,hou2019learning,Wang2020UnderstandingCR}, restricting series and image representations in the hypersphere makes sure the learned mixed representation $m_\lambda(\mathbf{u},\mathbf{v})$ contains both numerical and structural information of time series.
Our geodesic mixup strategy ensures the mixed representations remain on the unit hypersphere between two representations because $||m_\lambda(\mathbf{u},\mathbf{v})||=1$.
As shown in Fig. \ref{fig:mixup}(a), we treat these mixed representations as negative samples, such negative samples consider both numerical and structural patterns of time series data, thereby distinguishing time series that belong to different classes.
Positive samples retain the original series-image loss as used in Eq. \ref{eq:si}, giving rise to a new geodesic mixup contrastive loss as:
\begin{equation} \begin{split}     \ell_i^{\mathrm{I-mix}}=-\mathrm{log}\frac{\exp\bigl(\text{sim}(\mathbf{u}_{i}\cdot\mathbf{v}_{i})/\tau\bigr)}{\sum_{j=1}^B\exp\bigl(\text{sim}(\mathbf{u}_{i}\cdot m_\lambda(\mathbf{u}_{j},\mathbf{v}_{j}))/\tau\bigr)}\\     \ell_{i}^{\mathrm{S-mix}}=-\mathrm{log}\frac{\exp\bigl(\text{sim}(\mathbf{v}_{i}\cdot\mathbf{u}_{i})/\tau\bigr)}{\sum_{j=1}^{B}\exp\bigl(\text{sim}(\mathbf{v}_{i}\cdot m_\lambda(\mathbf{u}_{j},\mathbf{v}_{j}))/\tau\bigr)}, \end{split} \end{equation}  
where $\ell_i^{\mathrm{I-mix}}$  is the $i$-th image representation contrast with all representations after combing in one batch, and $\ell_i^{\mathrm{S-mix}}$ is the $i$-th time series representation contrast with all representations after combing. The geodesic mixup contrastive loss is defined as:
\begin{equation}     \mathcal{L}_{\mathrm{mix}}=\frac{1}{2B}\sum_{i=1}^{B}\left(\ell_{i}^{\mathrm{I-mix}}+\ell_{i}^{\mathrm{S-mix}}\right). \end{equation}
By summing the two losses, we obtain a combined loss for training in series-image contrastive learning as:
\begin{equation}     \mathcal{L}_{\mathrm{SI}}= \beta\mathcal{L}_{\mathrm{naive}}+(1-\beta)\mathcal{L}_{\mathrm{mix}},\end{equation}
where $\beta$ is a hyperparameter.

\begin{table*}[t]
  \centering
  \caption{Comparison with state-of-the-art representation learning methods in the case-by-case paradigm.}
  \fontsize{9pt}{10pt}\selectfont
  \setlength{\tabcolsep}{1mm}
    \begin{tabular}{cc|cccccccccc}
    \toprule
    \multicolumn{2}{c}{Method} & \multicolumn{1}{c}{AimTS} & \multicolumn{1}{c}{TimesURL} & \multicolumn{1}{c}{Data2Vec} & \multicolumn{1}{c}{InfoTS} & \multicolumn{1}{c}{TS2Vec} & \multicolumn{1}{c}
    {T-Loss} & \multicolumn{1}{c}{TNC}& \multicolumn{1}{c}
    {TS-TCC} \\
    \midrule
    \multirow{3}{*}{125 UCR datasets}&Avg.Acc&\textbf{0.870 } & 0.845  & 0.832  & 0.838  & 0.830  & 0.806  & 0.761  & 0.757  \\
    &Avg. Rank&\textbf{2.176} & 3.092 & 3.892  & 3.428  & 4.440  & 5.732  & 6.584  & 6.656  \\
    & Num.Top-1&\textbf{63} & 8     & 4     & 15    & 0     & 0     & 0     & 0 \\
    
    \midrule
    \multirow{3}{*}{30 UEA datasets} & Avg. ACC & \textbf{0.780}  & 0.752  & 0.738  & 0.714  & 0.704  & 0.658  & 0.670  & 0.668  \\
     & Avg. Rank & \textbf{1.967}  & 2.617  & 3.250  & 4.583  & 4.950  & 5.917  & 6.100  & 6.617  \\
     & Num.Top-1 & \textbf{13}    & 5     & 4     & 1     & 2     & 0     & 0     & 0 \\
      
    \bottomrule
  \end{tabular}
  \label{tab:allres}
\end{table*}

\begin{figure*}[t]
    \centering
    \begin{minipage}[t]{0.49\textwidth}
        \centering
        \includegraphics[width=\textwidth]{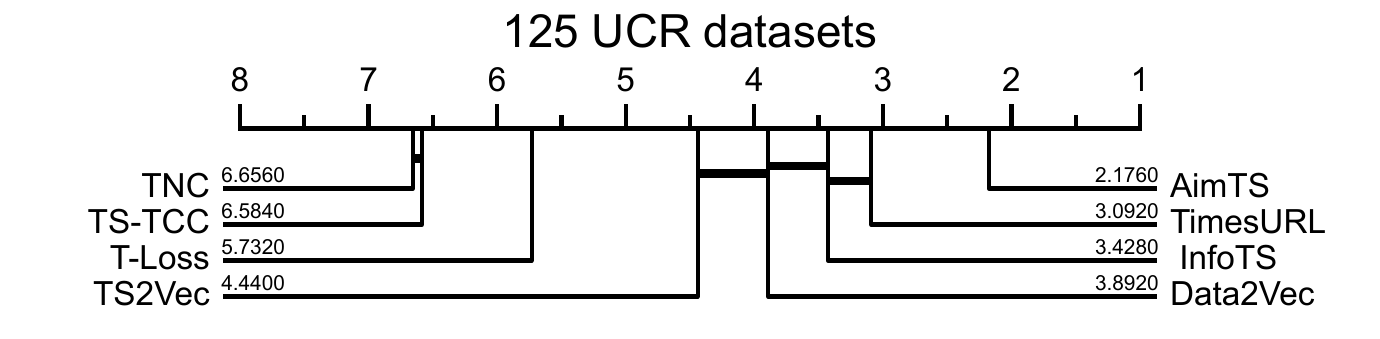}
    \end{minipage}
    \hfill
    \begin{minipage}[t]{0.49\textwidth}
        \centering
        \includegraphics[width=\textwidth]{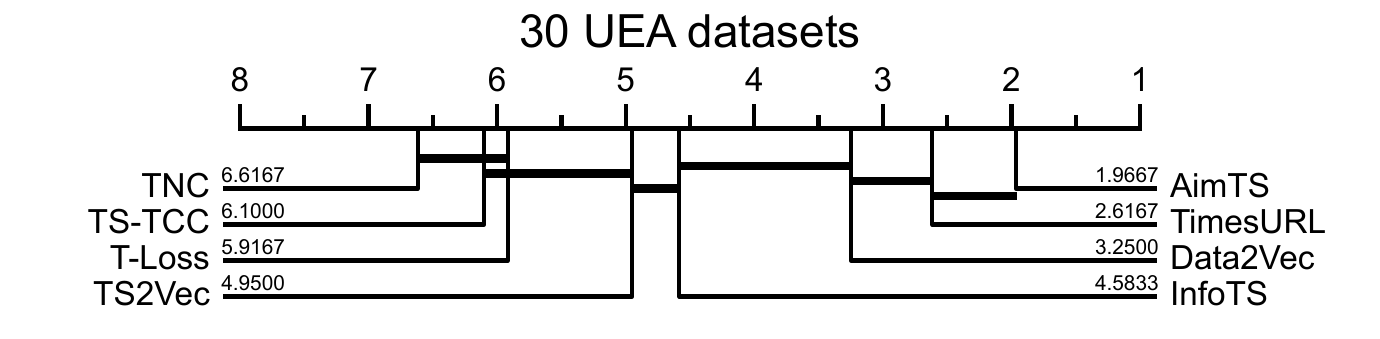}
    \end{minipage}
\caption{CD diagram of representation learning methods on UCR and UEA
datasets with a confidence level of 95\%.}
\label{cd}
\end{figure*}

\section{Experiments} \label{Experiments}

\subsection{Experimental Setup} \label{setting}
\subsubsection{Datasets}
\paragraph{Pre-training datasets}
\textbf{The Monash archive}~\cite{monash} includes 19 unlabeled datasets, of which 4 are univariate and 15 are multivariate. These datasets span various domains and contain between 24 and 4000 observations. 

\paragraph{Target datasets}
\textbf{The UCR archive}~\cite{ucr} consists of 128 univariate datasets in different domains, which are labeled with corresponding categories. \textbf{The UEA archive}~\cite{uea} contains 30 multivariate datasets.
We evaluated the performance of AimTS on downstream tasks using 158 datasets from these two archives, as well as the following datasets: \textbf{SleepEEG}~\cite{sleep}, \textbf{Epilepsy}\cite{Epilepsy}, \textbf{FD-B}\cite{FDB}, \textbf{Gesture}\cite{gesture}, and \textbf{EMG}\cite{EMG}. The training set of each dataset is used to fine-tune the pre-trained parameters of AimTS and train a classifier, which is then tested on the test set.

\paragraph{Few-shot learning datasets} Following UniTS \cite{units}, we perform few-shot learning on 6 datasets. ECG200 and StarLightCurves are from the UCR archive. Epilepsy, Handwriting, RacketSports and SelfRegulationSCP1 are from the UEA archive. We fine-tune AimTS and other baselines using 5\%, 15\% and 20\% of the training set from these 6 downstream datasets and evaluate their performance on the test set, respectively.

\subsubsection{Baselines} We compare AimTS with 29 baseline approaches across three paradigms. The case-by-case paradigm includes representation learning (e.g., TS-TCC~\cite{tstcc}, TS2Vec~\cite{TS2Vec}, Data2Vec~\cite{data2vec}), time series analysis method (e.g., TEST~\cite{sun2023test}, PatchTST~\cite{patchtst}, TimesNet \cite{timesnet}), and time series classification method (e.g., OS-CNN~\cite{oscnn}, TapNet~\cite{tapnet}, Rocket \cite{rocket}). All baselines are trained in a case-by-case setting. 

Models of the single source generalization paradigm (e.g., TF-C~\cite{tfc}, SimMTM~\cite{sim-mtm}, SoftCLT \cite{softcl}) often rely on labeled data beyond UEA and UCR for classification tasks. These methods are typically pre-trained on the SleepEEG dataset \cite{sleep} or Epilepsy dataset \cite{Epilepsy}, followed by fine-tuning using the training sets of Epilepsy dataset \cite{Epilepsy}, FD-B dataset\cite{FDB}, Gesture dataset \cite{gesture}, and EMG dataset\cite{EMG}, and finally evaluated on their test set, respectively. 
Since different methods are influenced by their pre-training datasets, making unified evaluation challenging, we use the best results reported in their papers as the baselines.
Multi-source adaptation foundation models are available for TSC. MOMENT~\cite{moment} collects multiple time series for multi-source pre-training from 4 task-specific, widely-used public repositories, including the UCR and UEA archives. UniTS \cite{units} pre-trains on 38 datasets from several sources, including 20 forecasting datasets and 18 classification datasets from UEA and UCR archives. Refer to MOMENT~\cite{moment} and UniTS \cite{units} for details.
We evaluate them using the full UCR and UEA archives.

\begin{table*}[h]
  \centering
  \caption{Comparison with other state-of-the-art methods in the case-by-case paradigm on 10 UEA datasets.}
  \fontsize{9pt}{10pt}\selectfont
  \setlength{\tabcolsep}{1mm}
    \begin{tabular}{l|cccccccccc}
    \toprule
    \multicolumn{1}{l|}{Method} & AimTS & TEST  & PatchTST & Crossformer & DLinear & TimesNet & OS-CNN & TapNet & Minirocket & Rocket \\
    \midrule
    EthanolConcentration & \textbf{0.563 } & 0.333  & 0.328  & 0.380  & 0.362  & 0.357  & 0.240  & 0.323  & 0.468& 0.447  \\
    FaceDetection & 0.677  & 0.581  & 0.683  & 0.687  & 0.680  & 0.686  & 0.575  & 0.556  & 0.620  & \textbf{0.694 } \\
    Handwriting & 0.482  & 0.414  & 0.296  & 0.288  & 0.270  & 0.321  & \textbf{0.668 } & 0.357  & 0.507  & 0.567  \\
    Heartbeat & \textbf{0.810 } & 0.725  & 0.749  & 0.776  & 0.751  & 0.780  & 0.489  & 0.751  & 0.771  & 0.718  \\
    JapaneseVowels & 0.989  & 0.962  & 0.975  & \textbf{0.991 } & 0.962  & 0.984  & 0.991  & 0.965  & 0.989  & 0.965  \\
    PEMS-SF & 0.850  & 0.800  & 0.893  & 0.859  & 0.751  & \textbf{0.896 } & 0.760  & 0.751  & 0.522  & 0.856  \\
    SelfRegulationSCP1 & \textbf{0.928 } & 0.819  & 0.907  & 0.921  & 0.873  & 0.918  & 0.835  & 0.652  & 0.925  & 0.866  \\
    SelfRegulationSCP2 & 0.578  & \textbf{0.591 } & 0.578  & 0.583  & 0.505  & 0.572  & 0.532  & 0.550  & 0.522  & 0.514  \\
    SpokenArabicDigits & 0.996  & 0.994  & 0.983  & 0.979  & 0.814  & 0.990  & \textbf{0.997 } & 0.983  & 0.620  & 0.630  \\
    UWaveGestureLibrary & \textbf{0.953 } & 0.885  & 0.858  & 0.853  & 0.821  & 0.853  & 0.927  & 0.894  & 0.938  & 0.944  \\
    \midrule
    Avg. ACC & \textbf{0.783} & 0.710  & 0.725  & 0.732  & 0.679  & 0.736  & 0.701  & 0.678  & 0.688  & 0.720  \\
    Avg. Rank & \textbf{2.800} & 6.250  & 5.600  & 4.300  & 7.750  & 4.550  & 5.850  & 7.300  & 4.550  & 5.350  \\
    Num.Top-1 & \textbf{4} & 1     & 0     & 1     & 0     & 1     & 2     & 0     & 0     & 1 \\
    \bottomrule
    \end{tabular}%
  \label{tab:uea10}%
\end{table*}%


\begin{table*}[h]
  \centering
  \caption{Compared with the state-of-the-art methods in the single source generalization paradigm.}
  \fontsize{9pt}{10pt}\selectfont
  \setlength{\tabcolsep}{1mm}
    \begin{tabular}{l|cccccccccccccc}
    \toprule
    Method & AimTS & SoftCLT & SimMTM & Ti-MAE & TST   & LaST  & TF-C  & CoST  & TS2Vec & SimCLR & TS-TCC & Mixing-up & CLOCS & TS-SD \\
\cmidrule{1-15}   EPILEPSY & \textbf{0.984 } & 0.970  & 0.955  & 0.803& 0.829  & 0.921& 0.950  & 0.937  & 0.945  & 0.907  & 0.925  & 0.802  & 0.951  & 0.895  \\
    FD-B  & \textbf{1.000 } & 0.805  & 0.694  & 0.680  & 0.656  & 0.467  & 0.694  & 0.548  & 0.607  & 0.492  & 0.550  & 0.679  & 0.493  & 0.557  \\
    GESTURE & 0.792  & \textbf{0.950 } & 0.800  & 0.755  & 0.751  & 0.642  & 0.764  & 0.733  & 0.733  & 0.480  & 0.719  & 0.693  & 0.443  & 0.692  \\
    EMG   & \textbf{1.000 } & \textbf{1.000 } & 0.976  & 0.635  & 0.759  & 0.663  & 0.817  & 0.732  & 0.809  & 0.615  & 0.789  & 0.302  & 0.699  & 0.461  \\
    \midrule
    Avg. ACC & \textbf{0.944} & 0.931  & 0.856  & 0.701  & 0.749  & 0.659  & 0.806  & 0.737  & 0.774  & 0.623  & 0.746  & 0.619  & 0.646  & 0.651  \\
    \bottomrule
    \end{tabular}%
  \label{tab:transfer}%
\end{table*}%

\subsubsection{Implementation details}

For pre-training, we implement AimTS in PyTorch~\cite{pytorch}, and all the experiments are conducted on 1 NVIDIA A800 80GB GPU. We use Adam~\cite{adam}  with an initial learning rate of $7 \times 10^{-3}$ and a random seed of 3407 for a batch size of 16 and implement learning rate decay using the StepLR method to implement learning rate decaying pre-training. 

After pre-training for 2 epochs, we can obtain the parameters of the TS encoder. We transfer the pre-trained model to each downstream task by fully fine-tuning~\cite{llm4ts} it and training an MLP as a classifier. By default, the optimizer uses Adam with a learning rate of 0.001 and the random seed is 3407. While obtaining the representations of the time series, we use channel independence~\cite{patchtst,shen} for the samples, encoding TS separately for each dimension of the time series.

\subsubsection{Data augmentation}
Following the previous work~\cite{augmentation1,infots,autocl}, we choose 5 data augmentations, including jittering, scaling, time warping, slicing, and window warping. 


\subsubsection{Evaluation metrics}
Following TS2Vec~\cite{TS2Vec}, we use several criteria that are considered important to evaluate classifiers, including the count of datasets achieving the highest accuracy \textit{(Num. Top-1)}, the average accuracy \textit{(Avg. ACC)}~\cite{acc}, the average ranking \textit{(Avg. Rank)}~\cite{cd} and \textit{Critical Difference (CD) diagram}~\cite{cd}.
\textit{Num. Top-1} shows the number of datasets where the model achieves the highest accuracy, excluding cases where more than one method shares the first place. 
\textit{Avg. ACC}~\cite{acc} is the average of the accuracy rates of multiple datasets and reflects the overall capability of the model, where higher values indicate better performance.
\textit{Avg. Rank} helps prevent the impact of extreme accuracy values on individual datasets, where lower values indicate better performance. 
\textit{CD diagram} uses statistical testing methods to more intuitively reflect the performance differences between different models. 
Models connected by a horizontal line indicate that they are not statistically different after the Friedman test.

\begin{table}[h]

  \centering
  \caption{Compared with the state-of-the-art methods in the multi-source adaptation paradigm.}
  \fontsize{9pt}{10pt}\selectfont
  \setlength{\tabcolsep}{1mm}
    \begin{tabular}{cc|ccc}
    \toprule
    \multicolumn{2}{c|}{Method} & AimTS & MOMENT & UniTS \\
    \midrule
    \multirow{3}[0]{*}{128 UCR datasets} & Avg. ACC & \textbf{0.870}  & 0.743  & 0.646  \\
          & Avg. Rank & \textbf{1.109}  & 2.172  & 2.719  \\
          & Num.Top-1 & \textbf{115}   & 6     & 2 \\
    \midrule
    \multirow{3}[0]{*}{30 UEA datasets} & Avg. ACC & \textbf{0.780}  & 0.696  & 0.639  \\
          & Avg. Rank & \textbf{1.083}  & 2.150  & 2.767  \\
          & Num.Top-1 & \textbf{26}    & 0     & 1 \\
    \bottomrule
    \end{tabular}%
  
  \label{tab:foundation-main}%
\end{table}%

\begin{table*}[t]
  \centering
  \caption{Few-shot learning on 6 downstream datasets.}
  \fontsize{9pt}{10pt}\selectfont
  \setlength{\tabcolsep}{1mm}
    \begin{tabular}{l|ccc|ccc|ccc}
    \toprule
    \multicolumn{1}{l|}{Data ratio} & \multicolumn{3}{c|}{5\% } & \multicolumn{3}{c|}{15\% }& \multicolumn{3}{c}{20\%} \\
    \multicolumn{1}{l|}{Method} & AimTS & MOMENT & UniTS & AimTS & MOMENT & UniTS & AimTS & MOMENT & UniTS \\
    \midrule
    ECG200 & \textbf{0.830 } & 0.640  & 0.790  & \textbf{0.850 } & 0.820  & 0.820  & 0.840 & \textbf{0.850}  & 0.820  \\
    StarLightCurves & \textbf{0.868 } & 0.791  & 0.826  & \textbf{0.931 } & 0.878  & 0.834  & \textbf{0.966 } & 0.950  & 0.833  \\
    Epilepsy & \textbf{0.848 } & 0.667  & 0.522  & \textbf{0.971 } & 0.833  & 0.681  & \textbf{0.949 } & 0.870  & 0.855  \\
    Handwriting & \textbf{0.193 } & 0.081  & 0.061  & \textbf{0.213 } & 0.081  & 0.080  & \textbf{0.233 } & 0.093  & 0.081  \\
    RacketSports & \textbf{0.533 } & 0.414  & 0.487  & \textbf{0.711 } & 0.513  & 0.618  & \textbf{0.743 } & 0.566  & 0.586  \\
    SelfRegulationSCP1 & \textbf{0.765 } & 0.706  & 0.758  & \textbf{0.850 } & 0.843  & 0.672  & \textbf{0.863 } & 0.862  & 0.737  \\
    \midrule
    Avg. ACC & \textbf{0.673 } & 0.550  & 0.574  & \textbf{0.754 } & 0.661  & 0.618  & \textbf{0.766 } & 0.699  & 0.652  \\
    \bottomrule
    \end{tabular}%
    
  \label{tab:fewshot}%
\end{table*}%

\subsection{Main Results}
\subsubsection{Compared to the case-by-case paradigm}
To illustrate that the learned representations of AimTS can be generalized to different classification tasks, we compare it with recently proposed representation learning methods for time series and report the results in Tab. \ref{tab:allres}. Furthermore, we show the CD diagrams with $\alpha=0.05$ of the Nemenyi test for all datasets in Fig. \ref{cd}, demonstrating that AimTS achieves the best overall average rankings in both the UCR archive and the UEA archive, which is higher than that of the existing representation learning methods. Notably, AimTS significantly outperforms these methods on the UCR dataset. 

Following TimesNet~\cite{timesnet} and conducted experiments on 10 UEA datasets to compare the performance of AimTS with existing supervised methods of the case-by-case paradigm. 
Although Num. Top-1 can reflect the model's performance to some extent, if it excels in this single metric but falls short in average metrics, it indicates that the method may only be effective on specific datasets.
On the 10 datasets, Avg. ACC of AimTS is 0.764, 2.8\% higher than the second-place TimesNet accuracy of 0.736.
In addition, AimTS has Avg. Rank of 2.8 on the 10 datasets, outperforming the second-placed Crossformer with Avg. Rank of 4.3 by 1.5.
These average metrics reflect the universal ability of our model across different datasets. 

\subsubsection{Compared to the single source generalization paradigm} To further demonstrate the generalizability of the representations learned by AimTS in multi-source pre-training, we compared AimTS with existing methods in the single source generalization paradigm on 4 datasets, shown as Tab. \ref{tab:transfer}. Due to the gap between the pre-training and fine-tuning datasets, the baselines perform poorly in most tasks. AimTS outperforms other baselines on the majority of datasets. Notably, for FD-B, AimTS significantly surpasses the previous state-of-the-art SoftCLT, with an accuracy of 1. These results demonstrate that AimTS effectively captures valuable knowledge during multi-source pre-training and achieves strong classification performance across various downstream tasks.

\subsubsection{Compared to the multi-source adaptation paradigm}
To comprehensively compare various paradigms, we also include time series foundation models in the evaluation. As shown in Tab. \ref{tab:foundation-main}, AimTS achieves the best results in 115 out of 128 datasets of the UCR archive and 26 out of 30 datasets of the UEA archive. In addition, it improves the classification accuracy by 12.7\% of 128 UCR datasets and 8.4\% of 30 UEA datasets, on average, over the second-best baseline.

Comprehensive comparisons with multiple baselines from the three paradigms demonstrate the effectiveness of AimTS.

\subsection{Few-Shot Learning}
As a pre-training model, AimTS demonstrates competitive few-shot capabilities. In this section, we compare AimTS with other pre-training models capable of few-shot learning on six downstream datasets, including UniTS~\cite{units} and MOMENT~\cite{moment}. We report the detailed results for each dataset in Tab. \ref{tab:fewshot}.

Compared to other foundation models, AimTS achieves outstanding performance with only 5\% of the data, nearly matching the classification accuracy that other baselines achieve with 15\% of the data. Furthermore, AimTS surpasses all baselines across all data ratios, achieving the highest average accuracy in every case. These experimental results highlight the outstanding generalization ability of AimTS, which maintains superior performance even under data-scarce conditions.

\begin{table}[h]
\centering
\caption{Ablation study of AimTS on 128 UCR datasets.}
\fontsize{9pt}{10pt}\selectfont
  \setlength{\tabcolsep}{1mm}
\begin{tabular}{cc|c}
    \toprule
       &  & Avg. Acc \\
        \midrule
    \multicolumn{2}{c|}{AimTS}
     & \textbf{0.870}\\

    \multicolumn{2}{c|}{w/ inter-prototype contrastive learning}
     &0.851\\

    \multicolumn{2}{c|}{w/ prototype-based contrastive learning}
     &0.858\\

     \multicolumn{2}{c|}{w/ naive series-image contrastive learning}
     &0.858\\

     \multicolumn{2}{c|}{w/ series-image contrastive learning}
     &0.865\\
     
    \bottomrule
    \end{tabular}
\label{tab:ablation1}
\end{table}

\subsection{Ablation Studies}
To verify the effectiveness of each component in AimTS, we decompose it into four parts based on the key contributions of the paper. A unified experimental setup is adopted: pre-training on the Monash dataset followed by testing on 128 downstream UCR datasets. The results are shown in the Tab. \ref{tab:ablation1}.

\subsubsection{Effect of inter-prototype contrastive learning}
We first validate the effectiveness of inter-prototype contrastive learning in Tab. \ref{tab:ablation1}. In this experiment, prototypes are obtained by simply averaging representations from different augmentations, and contrastive learning is applied between these prototypes. The results show that this approach achieved remarkable performance, even surpassing many baselines, demonstrating the necessity of using diverse augmentations and the validity of the prototype use.

\subsubsection{Effect of prototype-based contrastive learning}
We train using the complete two-level prototype-based contrastive learning, which includes both inter-prototype contrastive loss and intra-prototype contrastive loss. The results are shown in the second row of the Tab. \ref{tab:ablation1}. Compared to training with only inter-prototype contrastive learning, the complete method achieves 0.858, confirming the necessity of adjusting the distribution of augmented representations.

\subsubsection{Effect of naive series-image contrastive learning}
To validate the critical role of the image modality in AimTS, we train the model using only the series-image contrastive loss, achieving an accuracy of 0.858 on UCR, as shown in Tab. \ref{tab:ablation1}. This result also surpasses most baselines, further demonstrating the effectiveness of our approach.

\subsubsection{Effect of geodesic mixup strategy}
To validate the effectiveness of the geodesic series-image mixup strategy, we perform pre-training using the complete series-image contrastive loss, which combines the naive series-image contrastive loss and the geodesic mixup contrastive loss. As shown in Tab. \ref{tab:ablation1}, incorporating mixed samples into contrastive learning further improved the model's performance to 0.865.

\begin{figure*}[h]
    \centering
    \begin{minipage}[t]{0.22\textwidth}
        \centering
        \includegraphics[width=\textwidth]{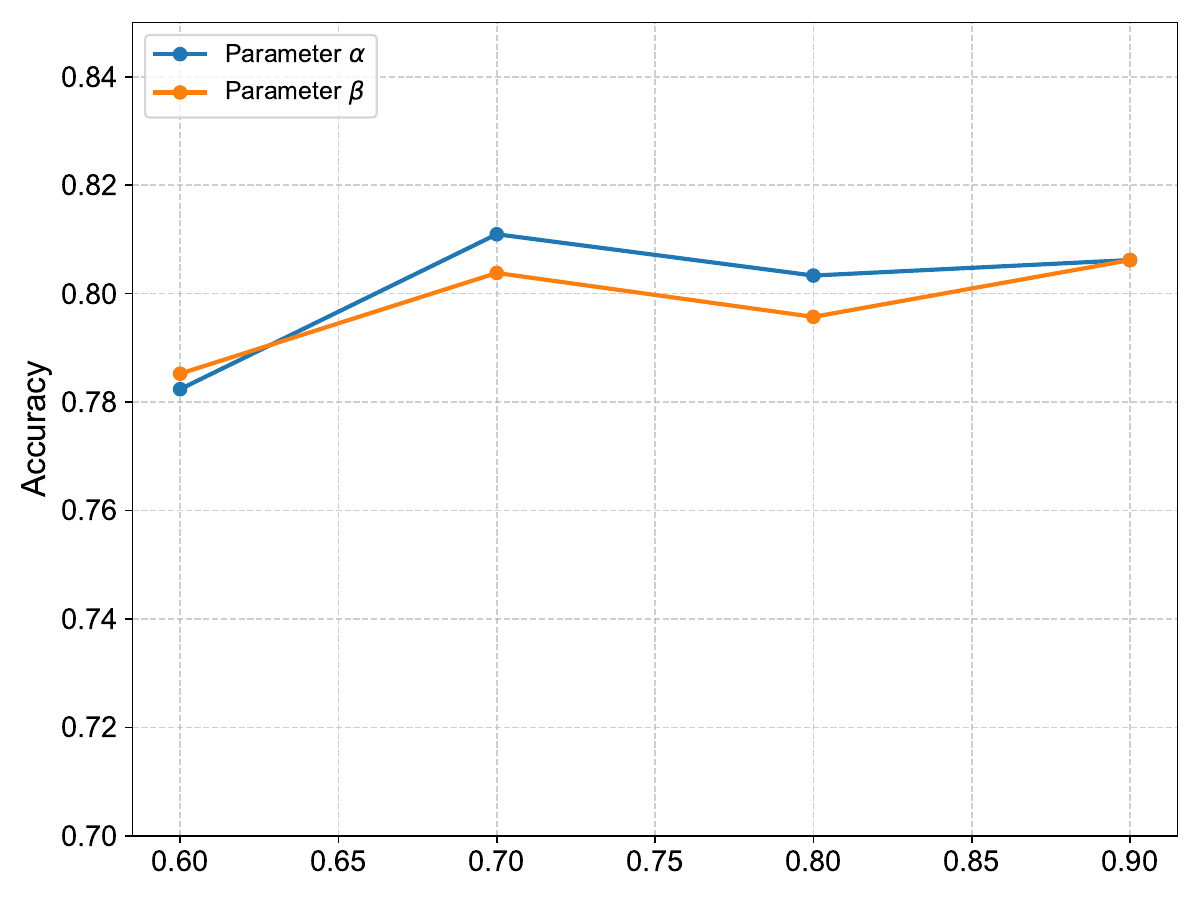}
        \subcaption{Effects of $\alpha$ and $\beta$.}
        \label{fig:paramab}
    \end{minipage}
    \hfill
    \begin{minipage}[t]{0.22\textwidth}
        \centering
        \includegraphics[width=\textwidth]{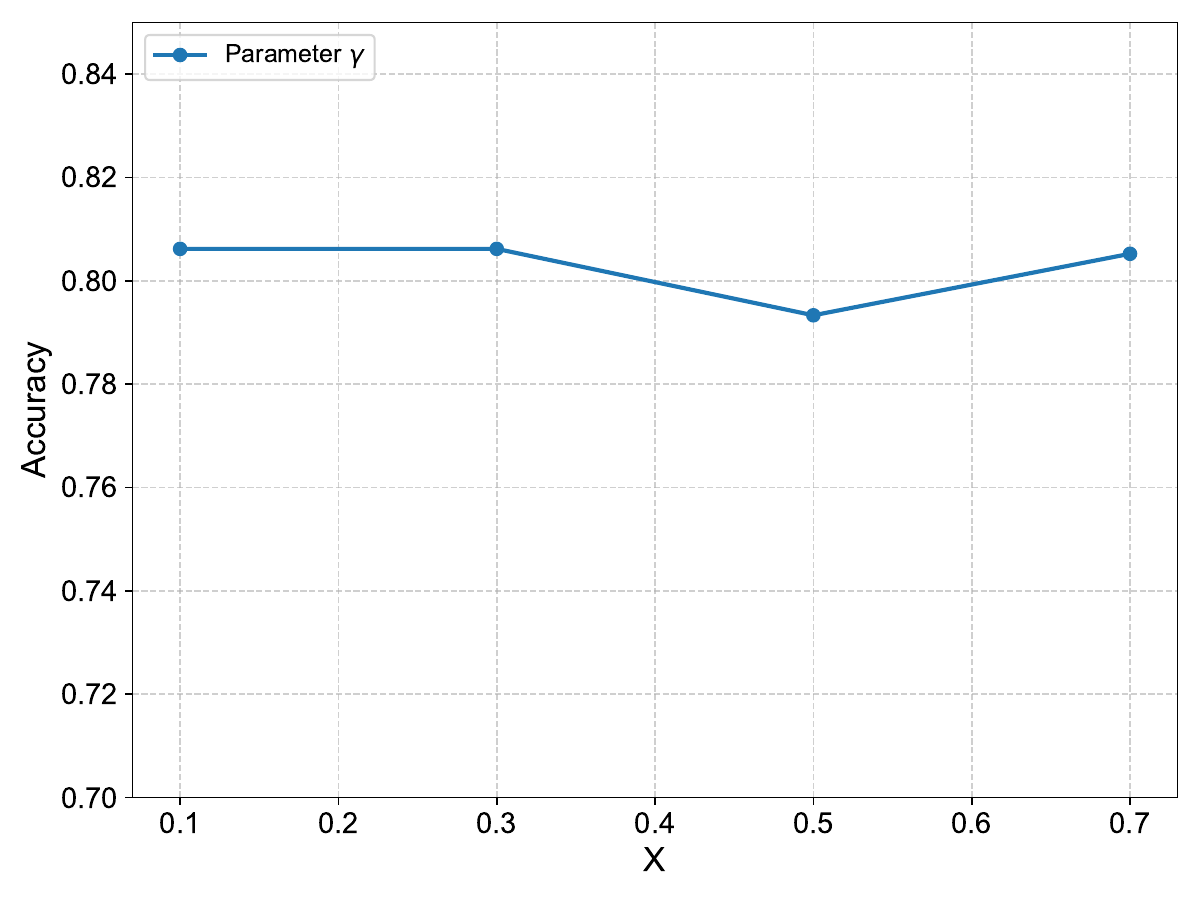}
        \subcaption{Effects of $\gamma$.}
        \label{fig:paramv}
    \end{minipage}
    \hfill
    \begin{minipage}[t]{0.22\textwidth}
        \centering
        \includegraphics[width=\textwidth]{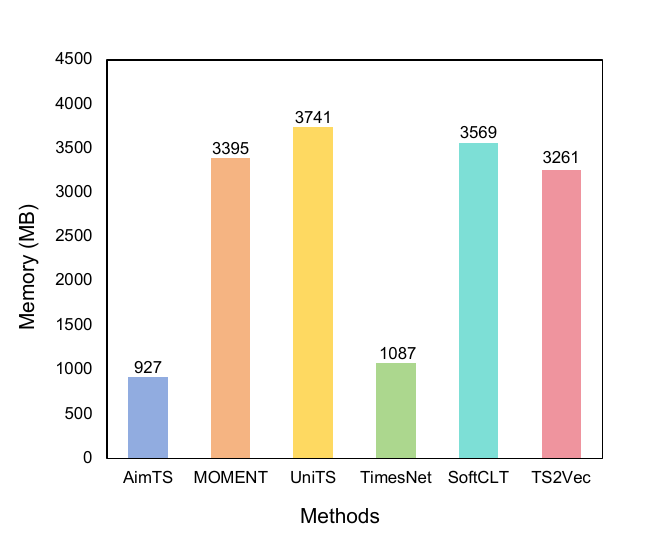}
        \subcaption{Memory comparison.}
        \label{fig:memory}
    \end{minipage}
    \hfill
    \begin{minipage}[t]{0.22\textwidth}
        \centering
        \includegraphics[width=\textwidth]{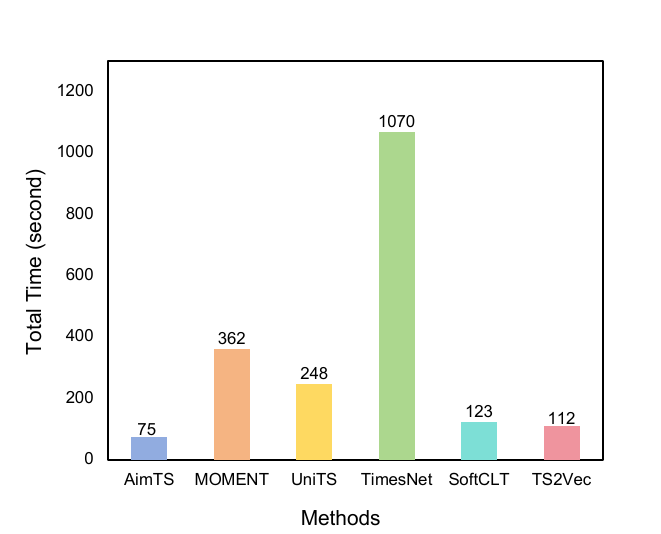}
        \subcaption{Efficiency comparison.}
        \label{fig:time}
    \end{minipage}
    
    \caption{(a)(b) Results of AimTS with different parameters. (c)(d) GPU memory usage and efficiency comparison on StarLightCurves.}
    \label{fig:more}
\end{figure*}

\subsection{Parameter Studies} 
To analyze the contribution of the proposed loss functions in our work, we conduct experiments on the weight hyperparameters associated with each loss. Additionally, in the time-series image contrast, the mixup coefficient $\lambda$ is sampled from a beta distribution $\mathrm{Beta}(\gamma, \gamma)$, where $\gamma$ is a hyperparameter. We explored the impact of $\gamma$ on the overall model performance. We select AllGestureWiimoteX, AllGestureWiimoteY, and AllGestureWiimoteZ datasets from UCR to conduct experiments. The sensitivity of AimTS to these parameters is evaluated based on average accuracy across these datasets.

\subsubsection{Effect of $\alpha$}
We examine the weight $\alpha$ for the intra-prototype contrastive loss in prototype-based contrastive learning. In this experiment, $\beta=0.1$ and $ \gamma=0.1$. We vary $\alpha$ between 0.9, 0.8, 0.7, and 0.6 during pre-training and evaluate the performance on downstream datasets. As shown in the Fig. \ref{fig:more}(a), $\alpha$ has a limited impact on the performance of AimTS. AimTS achieves the best accuracy when $\alpha=0.7$, while performance degrades slightly when $\alpha=0.6$. Notably, since the purpose of intra-prototype contrastive learning is to refine prototypes, we did not test values less than 0.5.

\subsubsection{Effect of $\beta$}
Similarly, we explore the influence of $\beta$ on the mixup contrastive loss in series-image contrastive learning. We fix the parameters $\alpha=0.1, \gamma=0.1$. $\beta$ is set to 0.9, 0.8, 0.7, and 0.6, with results shown in the Fig. \ref{fig:more}(a). To ensure accurate correspondence between time series and image representations, the series-image contrastive loss is given a consistently higher weight. The results indicate that $\beta$ has minimal impact, with the best performance observed when $\beta=0.9$. 

\subsubsection{Effect of $\gamma$}
In mix contrastive learning, the mixup coefficient $\lambda\sim\mathrm{Beta}(\gamma,\gamma)$ is a random coefficient to control the ratio of image and series modalities representations. $\gamma$ influences the shape of the beta distribution, typically ranging between 0 and 1. To investigate whether different forms of the beta distribution affect the mixup strategy, we conduct experiments by varying $\gamma$, as shown in Fig. \ref{fig:more}(b). The performance of AimTS remains stable with $\gamma$ set to 0.1, 0.3, 0.5, and 0.7 when parameters $\alpha=0.1,\beta=0.1$, demonstrating that this mixup strategy is not sensitive to the hyperparameter and is a generalizable method.

\begin{figure*}[t]
    \centering
    \begin{minipage}[t]{0.23\textwidth}
        \centering
        \includegraphics[width=\textwidth]{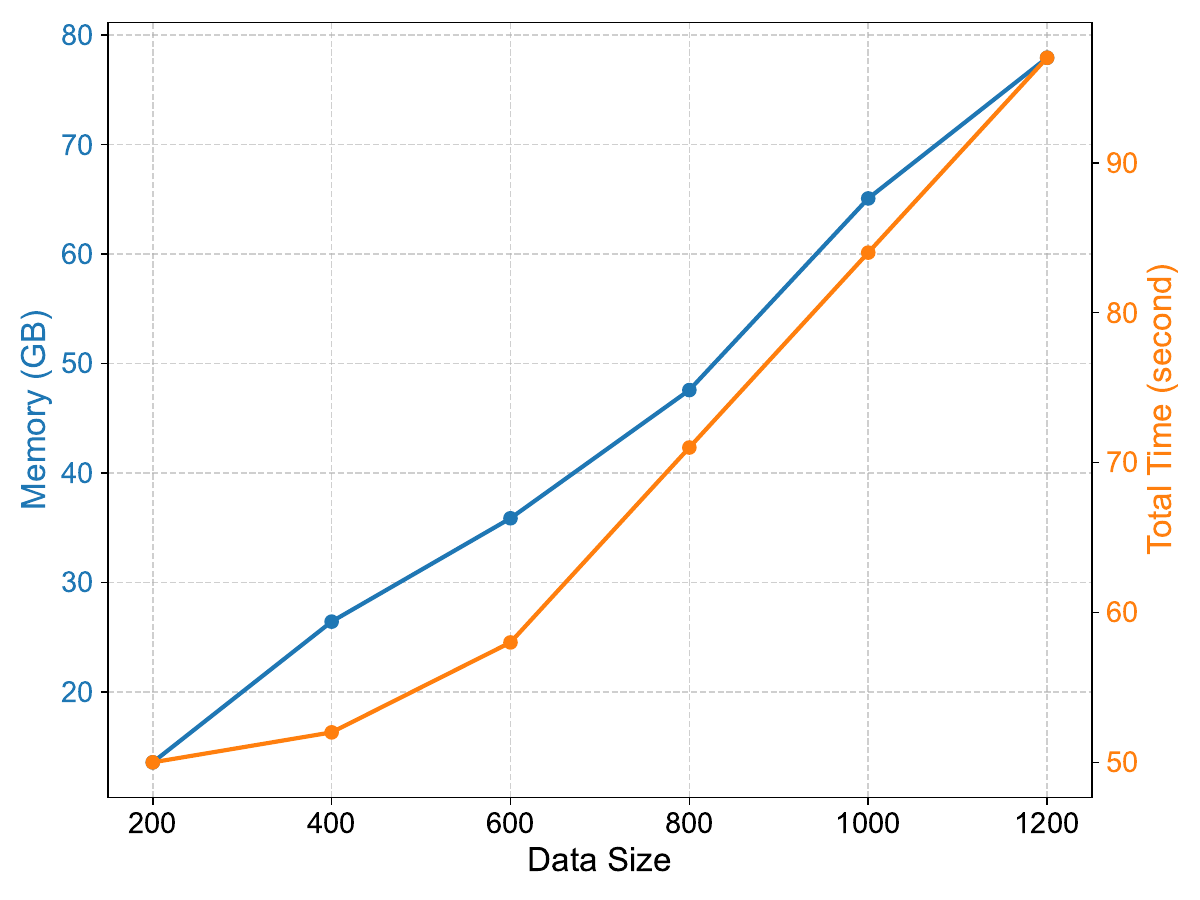}
        \subcaption{Memory and efficiency w.r.t. Data Size.}
        \label{fig:scale1}
    \end{minipage}
    \hfill
    \begin{minipage}[t]{0.23\textwidth}
        \centering
        \includegraphics[width=\textwidth]{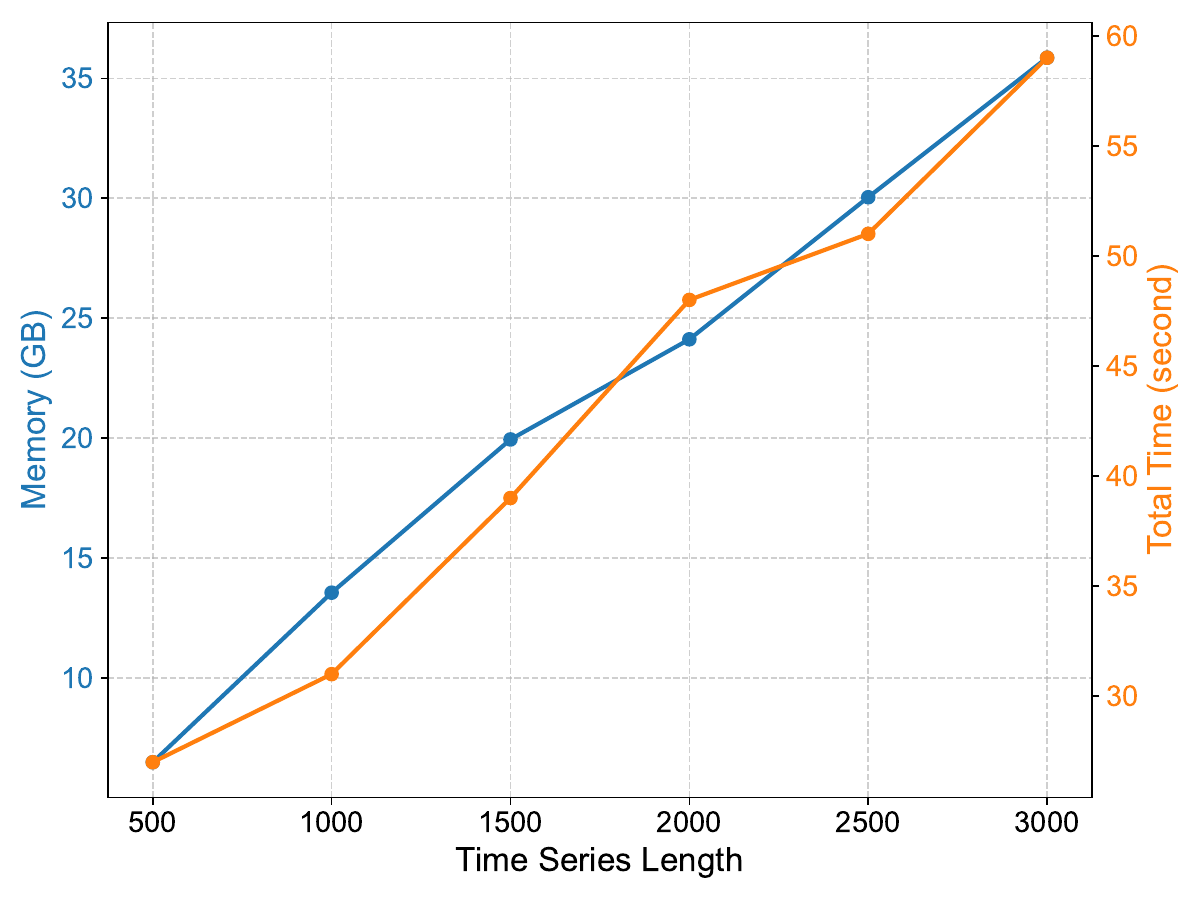}
        \subcaption{Memory and efficiency w.r.t. Time Series Length.}
        \label{fig:scale2}
    \end{minipage}
    \hfill
    \begin{minipage}[t]{0.23\textwidth}
        \centering
        \includegraphics[width=\textwidth]{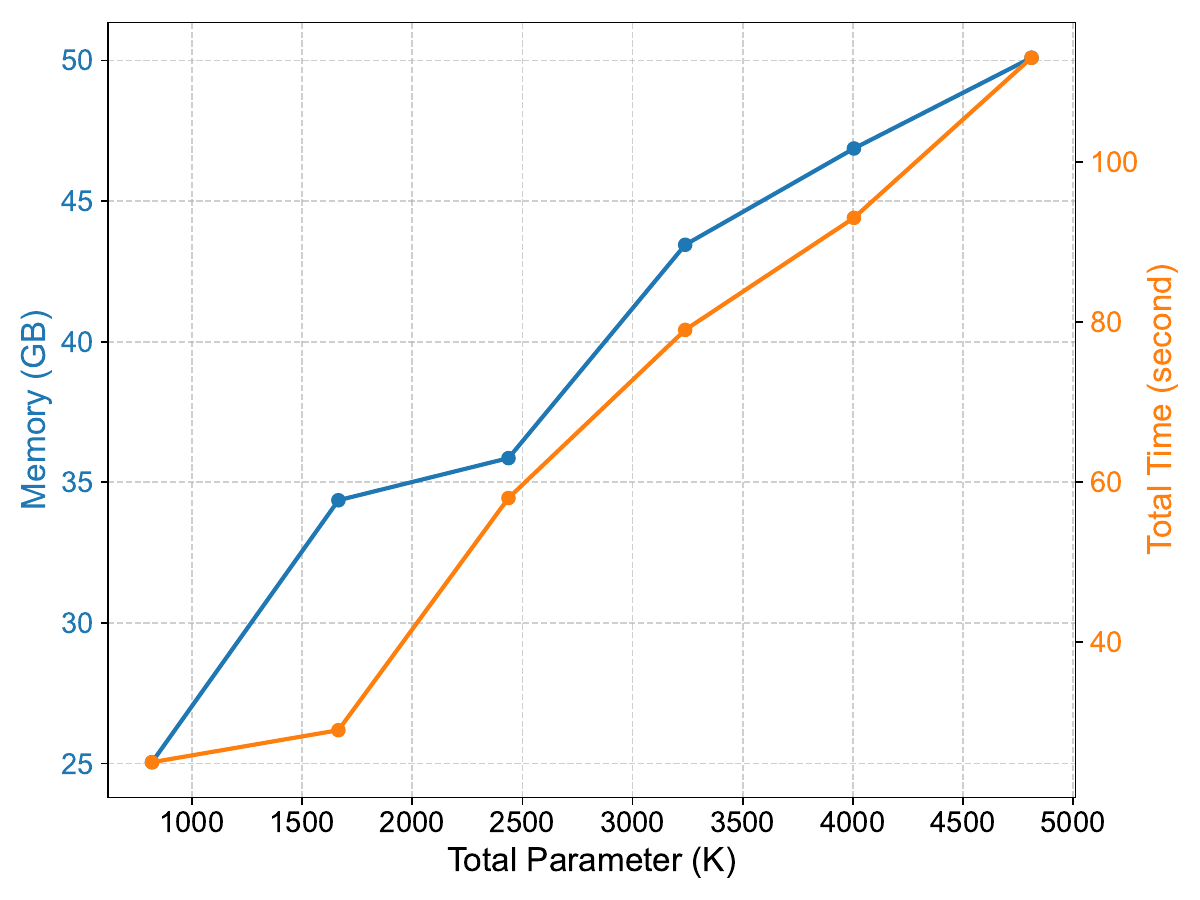}
        \subcaption{Memory and efficiency w.r.t. Model Parameter.}
        \label{fig:scale3}
    \end{minipage}
    \hfill
    \begin{minipage}[t]{0.23\textwidth}
        \centering
        \includegraphics[width=\textwidth]{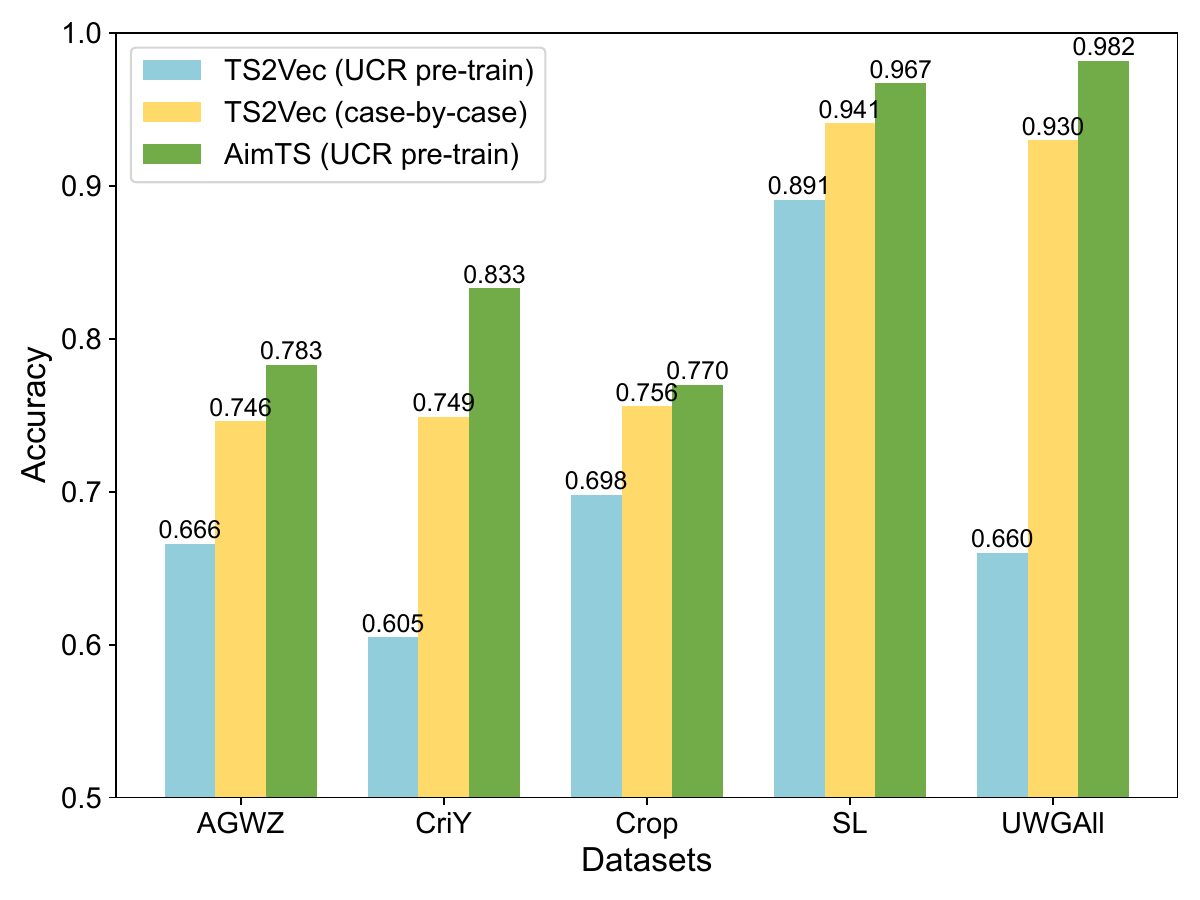}
        \subcaption{Challenge of pre-training.}
        \label{fig:challenge}
    \end{minipage}
    \caption{(a)(b)(c) Scalability comparison on SleepEEG dataset. (d) Results of TS2Vec in a case-by-case setting, TS2Vec pre-trained with a multiple domain dataset, and AimTS.}
    \label{fig:scale}
\end{figure*}

\subsection{Memory Usage and Efficiency}
We compare the GPU memory usage and efficiency of AimTS and 5 baselines on the StarLightCurves dataset. For AimTS, MOMENT~\cite{moment} and UniTS \cite{units}, we fine-tune the pre-trained parameters and train a classifier on the training set of the dataset. The parameters of other baselines are trained using the training set of the dataset. To ensure fairness, the batch size for all methods is 8, and the number of epochs is 10.

Fig. \ref{fig:more}(c) reports the maximum GPU memory usage of all methods during fine-tuning or training. Fig. \ref{fig:more}(d) reports the total time for fine-tuning or training, and inference for all methods. During fine-tuning and inference, AimTS requires only 927 MB of GPU memory, which is 14.72\% lower than the second baseline TimesNet. In addition to lower memory requirements, AimTS achieves a total time of 75 seconds, which is faster than other models. In summary, AimTS achieves superior efficiency, requiring significantly less memory and time without compromising performance.

\subsection{Scalability Studies} 
To evaluate the scalability of AimTS, we analyze the impact of three critical factors: dataset size, time series length, and model parameters on GPU memory usage and total time of fine-tuning and testing. Unlike other time series tasks, in time series classification, the length of the series does not affect the model parameters but instead impacts the size of the data processed in each batch. Therefore, the time series length is analyzed as a separate factor. All experiments are conducted on the SleepEEG dataset, with all settings kept consistent except for the subject under study. For each factor, we present detailed analyses supported by line plots, as shown in Fig. \ref{fig:scale}(a)(b)(c).

\subsubsection{Data size}
To evaluate the impact of data size on GPU memory usage and total running time, we fixed the time series length at 3000 and the model parameters at 2437K while increasing the amount of data used for fine-tuning. 
The GPU memory usage and running time scale linearly with the size of the fine-tuning dataset, as shown in Fig. \ref{fig:scale}(a). GPU memory usage increases steadily as data size grows, reflecting the demand for larger batches of data storage. Similarly, the total training time increases at a proportional rate due to the increased number of iterations required to process the larger dataset.

\subsubsection{Time series length} In this experiment, the fine-tuning data size is 600, and the total parameters are configured to 2437K. We recorded the maximum GPU memory usage and total time required for fine-tuning AimTS when classifying time series of varying lengths. As depicted in Fig. \ref{fig:scale}(b), both GPU memory usage and training time exhibit a linear increase with the length of the time series. This behavior is expected, as longer time series require proportional computational resources and memory allocation. Importantly, the linear scaling highlights the computational efficiency of AimTS when handling long time series, making it highly suitable for downstream tasks with large time series lengths.

\subsubsection{Model parameter} When analyzing the impact of model parameters, we fixed the data size at 600 and the time series length at 3000. The scalability of AimTS with respect to its parameter size is analyzed in Fig. \ref{fig:scale}(c). As expected, both memory usage and running time increase with the number of parameters, and the growth rate is moderate.

\subsection{Additional Analyses}
\subsubsection{Challenge of multi-source pre-training}
Due to the data coming from different domains, the semantic differences pose challenges for pre-training. This section presents experiments to demonstrate that previous methods struggle to handle such issues, while AimTS overcomes them. TSVec is used as the baseline, with TS2Vec and AimTS pre-trained on the training set of the UCR datasets, and fine-tuned on 5 downstream datasets. It can be observed in Fig. \ref{fig:scale}(d) that TS2Vec, when using the multi-source pre-training and fine-tuning paradigm, performs worse than the case-by-case paradigm, indicating negative transfer caused by multi-source pre-training. AimTS, using multi-source datasets, performs exceptionally well in downstream tasks, demonstrating its strong generalization. 

    
    


\begin{table}[h]
\centering
\caption{Pre-trained AimTS on different datasets.}
\fontsize{9pt}{10pt}\selectfont
  \setlength{\tabcolsep}{1mm}
    \begin{tabular}{c|ccc}
    \toprule
    Pre-train Data& \multicolumn{1}{c}{Monash} & \multicolumn{1}{c}{UCR} & \multicolumn{1}{c}{UEA} \\
    \midrule
    128 UCR datasets &       0.870& 0.871&  0.858\\
    30 UEA datasets &       0.780&   0.774&  0.782\\
    \bottomrule
    \end{tabular}
\label{tab:pretraindata} 
\end{table}
\subsubsection{Comparison of different datasets used for pre-training} \label{diffdata}
To validate that AimTS can obtain generalized representations through pre-training on different multi-source datasets, we conduct pre-training on various datasets. Pre-training AimTS using the UCR data indicates that we combined the training samples from 128 datasets into one pre-training dataset. 
We use the training data from the UEA archive for AimTS pre-training. Tab. \ref{tab:pretraindata} compares the average accuracy of AimTS pre-trained using three multi-source datasets. This result confirms that AimTS can obtain generalized representations across different multi-source datasets. Additionally, the results show that AimTS achieves better performance on downstream datasets when it has been exposed to these datasets during the pre-training, which reaffirms Paradigm 3 mentioned in the introduction as a more straightforward approach.

\subsubsection{Case study of semantic changes caused by data augmentation}

To demonstrate the motivation of prototype-based contrastive learning, we conduct a case study to show the phenomenon that data augmentation may change the semantics of the data.
\begin{figure}[h]
    \centering
    \includegraphics[width=1\linewidth]{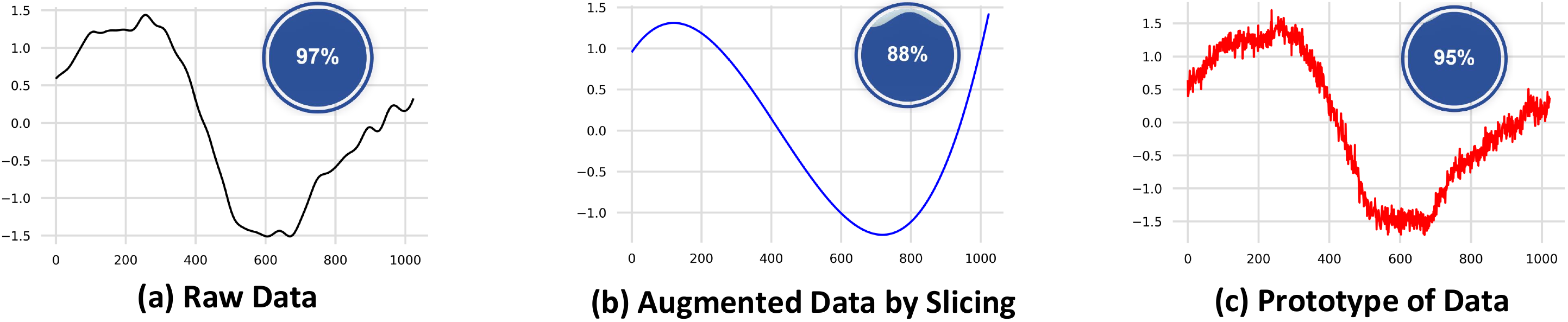}
    \caption{Test with different data.}
    \label{differentdata}
    \vspace{-15pt}
\end{figure}
Fig. \ref{differentdata} visualizes (a) a piece of raw time series data from the StarLightCurves dataset, (b) its augmented data using the slicing augmentation \cite{le2016data} that randomly crops the input time series and then linearly interpolates it back to the original length, and (c) its prototype generated using multiple augmentations, respectively.  
We train TS2Vec by the train dataset as a classifier and test whether the time series data still correspond to original labels, thereby assessing whether the augmentation influences the classification accuracy.
The top-right bubble in each sub-figure is the accuracy of the classifier.

The classifier achieves an accuracy of 0.97 on the raw test dataset, as shown in Fig. \ref{differentdata}(a). 
When testing on the augmented test dataset by slicing, the accuracy is 0.88 as shown in Fig. \ref{differentdata}(b). 
This indicates that slicing changes the semantics of many test data samples, causing them to no longer correspond to original labels. 
When testing on the prototypes of test data, the accuracy is 0.95 as shown in Fig. \ref{differentdata}(c), which is close to the accuracy of the raw dataset. 
In addition, in the three pieces in Fig. \ref{differentdata}, the classifier correctly classified raw data and prototype of data, but misclassified augmented data by slicing.
This shows that certain data augmentation methods may change semantic information while using prototypes helps maintain semantic consistency.

\section{Conclusion}
This paper presents AimTS, a multi-source pre-training framework designed to learn generalized representations and enhance various downstream time series classification tasks. 
AimTS proposes a two-level prototype-based contrastive learning method, effectively utilizing various augmentations and avoiding semantic confusion caused by augmentations in multi-source pre-training.
Considering augmentations within the time series modality are insufficient to address the classification problems with distribution shift, AimTS introduces image modality to capture structural information of time series data.
Experimentally, representations pre-trained by the AimTS can be fine-tuned for various classification tasks, and its performance outperforms the state-of-the-art methods while also demonstrating efficiency in terms of memory usage and computational costs.



\section*{Acknowledgments}
This work was supported by the National Natural Science Foundation of China (No. 62372179, 62406112).

\bibliographystyle{IEEEtran}
\bibliography{IEEEabrv,mybibfile}

\end{document}